\documentclass[journal]{IEEEtran}
\usepackage{cite}
\usepackage{amsmath,amssymb,amsfonts}
\usepackage{algorithmic}
\usepackage{graphicx}
\usepackage{textcomp}

\usepackage{bbding}
\usepackage{gensymb}
\usepackage{algorithm}
\usepackage{xcolor}
\usepackage{mathtools}
\usepackage{amsmath}
\usepackage{amssymb}
\usepackage{url}
\usepackage{array}
\usepackage{subfigure} 
\usepackage{stfloats}
\usepackage{multicol}
\usepackage[final]{pdfpages}

\ifCLASSINFOpdf
\else
\fi
\begin{document}
%
\title{Leveraging Regular Fundus Images for Training UWF Fundus Diagnosis Models via Adversarial Learning and Pseudo-Labeling}
\author{Lie Ju, Xin Wang, Xin Zhao, Paul Bonnington, Tom Drummond, and 
	Zongyuan Ge

\thanks{(Corresponding author: Zongyuan Ge)}
\thanks{Lie Ju and Zongyuan Ge are with Monash University, Clayton, VIC
3800 Australia, and also with Airdoc, Beijing 100000, China (E-mail:
julie334600@gmail.com, zongyuan.ge@monash.edu).}
\thanks{Xin Wang and Xin Zhao are with Airdoc, Beijing 100000, China (E-mail:
wangxin@airdoc.com, zhaoxin@airdoc.com).}
\thanks{Paul Bonnington and Tom Drummond are with Monash University, Clayton, VIC
3800 Australia (E-mail: Paul.Bonnington@monash.edu, Tom.Drummond@monash.edu).}
}

\maketitle

\begin{abstract}
Recently, ultra-widefield (UWF) 200\degree~fundus imaging by Optos cameras has gradually been introduced because of its broader insights for detecting more information on the fundus than regular 30$\degree$ - 60$\degree$ fundus cameras. Compared with UWF fundus images, regular fundus images contain a large amount of high-quality and well-annotated data.
Due to the domain gap, models trained by regular fundus images to recognize UWF fundus images perform poorly. Hence, given that annotating medical data is labor intensive and time consuming, in this paper, we explore how to leverage regular fundus images to improve the limited UWF fundus data and annotations for more efficient training.
We propose the use of a modified cycle generative adversarial network (CycleGAN) model to bridge the gap between regular and UWF fundus and generate additional UWF fundus images for training. A consistency regularization term is proposed in the loss of the GAN to improve and regulate the quality of the generated data. Our method does not require that images from the two domains be paired or even that the semantic labels be the same, which provides great convenience for data collection. Furthermore, we show that our method is robust to noise and errors introduced by the generated unlabeled data with the pseudo-labeling technique. We evaluated the effectiveness of our methods on several common fundus diseases and tasks, such as diabetic retinopathy (DR) classification, lesion detection and tessellated fundus segmentation. The experimental results demonstrate that our proposed method simultaneously achieves superior generalizability of the learned representations and performance improvements in multiple tasks.
\end{abstract}


\section{Introduction}
Retinal diseases are the main causes of blindness, with cataracts and glaucoma ranking first and second~\cite{fu2018joint,tham2014global}. Other diseases, such as diabetic retinopathy (DR) and age-related macular degeneration (AMD), also increase the risk of blindness in elderly individuals~\cite{kobrin2007overview,kocur2002visual,gulshan2016development}. Furthermore, high myopia was reported to be the second most frequent cause of blindness in China~\cite{xu2006causes}. Many people suffering from early-stage fundus diseases do not have vision loss, but once vision problems begin to occur, it is too late to intervene; this phenomenon demonstrates the deterioration related to the disease~\cite{nussenblatt2007age,centers2011national,shaw2010global}. Worst of all, vision loss from these diseases is always irreversible, so early screening and detection methods with specific equipment are vital and essential to preserve vision and quality of life.

\begin{figure}[!t]
	\centering
{\includegraphics[width=8cm]{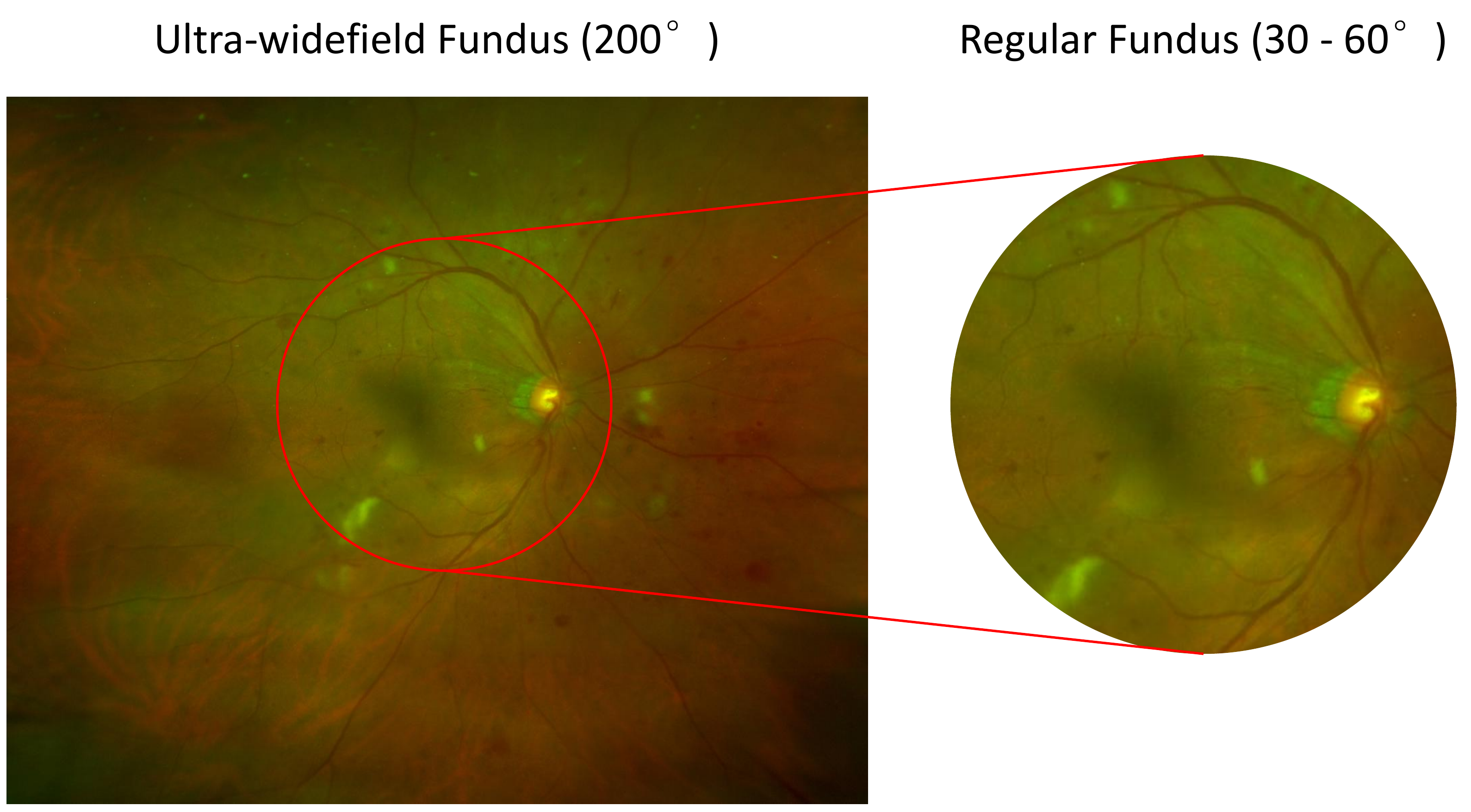}}
\caption{Comparison of regular fundus and the UWF fundus imaging views. Regular fundus imaging can only cover the red circle, which makes it difficult to detect lesions that are out of range. The fundus has actually met one of the 4-2-1 criteria, but it will be misdiagnosed as NPDRII in regular fundus imaging. 
} \label{fig1}
\end{figure}

Over the last 50 years, regular fundus camera for screening has been commonly used to detect abnormal retinal diseases. It can provide a relatively good prognosis of visual acuity in the early stage. However, although regular fundus screening can detect most retinal diseases, it still has its own limitations. For example, cataracts, vitreous opacity and other diseases characterized by weak refractive stroma are often difficult to image via traditional examinations because of obstruction on the optical path. Optos ultra-widefield (UWF) fundus imaging first became commercially available in the 2000s, and the image-capturing range can cover 80\% of the area/200\degree ~of the retina, compared to only 30\degree ~- 60\degree ~achieved with regular retinal cameras. UWF imaging is 
particularly well-suited to use in teleophthalmology, as it has a rapid acquisition time 
for the area of retina imaged, and allows better detection of peripheral retinal pathology. 
 As Fig.~\ref{fig1} shows, UWF imaging covers a greater retinal area than regular imaging, providing more clinically relevant information about the pathology, which usually changes from the peripheral retina to be detected, such as retinal degeneration, detachment, hemorrhages, and exudations. The clinical deployment of UWF imaging in the field of DR screening has demonstrated 
lower rates of ungradable images, reduced image evaluation time, and higher rates of 
pathology detection~\cite{li2018efficacy}.  \cite{kiss2014ultra,nagiel2016ultra} show that UWF fundus imaging provides new insights into a variety of disorders, even those that primarily involve the posterior pole. It further indicates that that diabetic retinopathy may need to take 2-3 times at different angles by the regular fundus camera to make a complete accurate diagnosis, while UWF only takes once.
Also, UWF imaging offers additional diagnostic information for 
vessel detection. \cite{ding2020weakly} study how to leverage extra UWF fluorescein 
angiography (FA) images for training. And \cite{tan2020deep} also show that DL-based analysis of UWF images has the potential to play an important role in the future 
delivery of teleophthalmology screening platforms, which is considered to have great clinical meaning and practical value.

For years, deep learning has been widely used on regular fundus images and has achieved good performance in the diagnosis of various retinal diseases.  \cite{fu2018joint, fu2018disc} proposed segmenting the optic disc and cup to help diagnose glaucoma based on the polar transformation and disc-aware ensemble network. \cite{jin2019dunet} used a deformable network named deformable U-Net (DUNet) for retinal vessel segmentation. \cite{wang2019two} proposed a two-stream convolutional neural network (CNN) specifically designed for multimodal AMD categorization. \cite{wang2019retinal} designed a multitask deep learning framework for identifying 36 retinal diseases. For UWF fundus image recognition, \cite{nagasato2018deep,ohsugi2017accuracy} recently used deep learning and support vector machine (SVM) algorithms to detect central retinal vein occlusion (CRVO) and rhegmatogenous retinal detachment (RRD). \cite{levenkova2017automatic} used hand-made features and SVM to detect DR features in the peripheral retina. \cite{nagasawa2019accuracy} investigated using UWF fundus images with a deep convolutional neural network (DCNN) to detect proliferative DR (PDR). \cite{li2020deep} developed a cascaded deep learning system for automated retinal detachment detection and macula-on/off retinal detachment discerning. \cite{ohsugi2017accuracy} evaluated the performance of DCNN and SVM in the detection of RRD. However, we find that there are still limitations of recent research studies about retinal disease diagnosis with UWF fundus images. First, most literature studies focus on only one specific disease and do not show the advantages of UWF fundus imaging compared with regular fundus imaging. For example, \cite{ohsugi2017accuracy} resized the image into 96 $\times$ 96 pixels as the input size. However, for most retinal detection tasks, the image with such small size will lose its important pathology and semantic information and the advantage of the high resolution of UWF fundus imaging is not fully utilized. Second, the dataset being trained and tested only contains clean and ideal samples collected from a controlled environment. In real scenarios, UWF fundus images are often obstructed by the eyelids and eyelashes. These artifacts may affect the screening performance of the model trained on clean images. Third, although the UWF screening equipment has been put into use gradually in recent years, the high-quality annotated data that can be used for deep learning training are still very scarce, most of which focus on one specific task, and the universality of the algorithm that can be applied to various complex retinal diseases and screening tasks is still missing.

\begin{figure}
	\includegraphics[width=9cm]{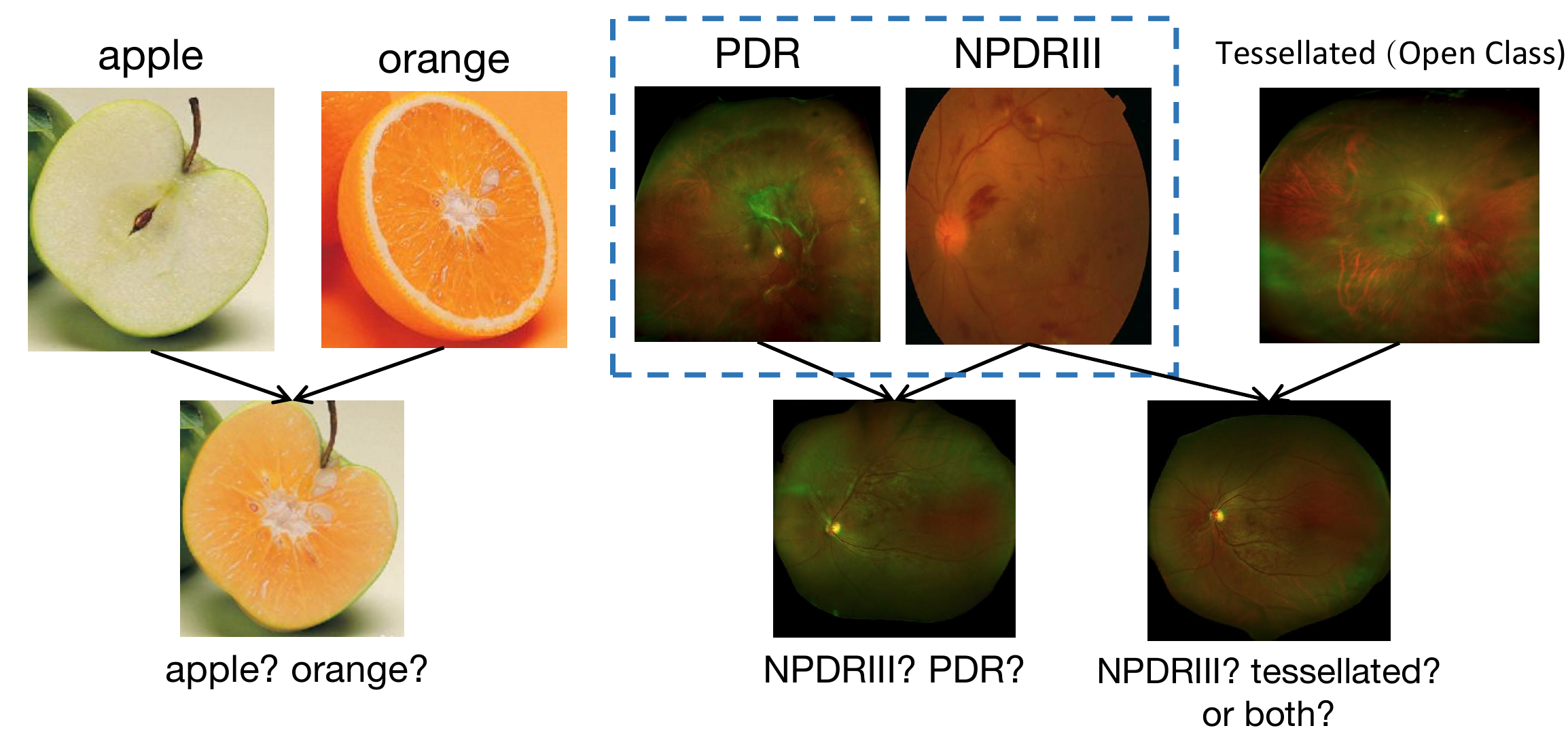}
	\centering
\caption{The figure shows the potential uncontrollability of GAN training. For simple tasks (apple and orange), a single or specific feature can be transferred well. However, for complex tasks, it is difficult to define the generated image, so the original label from the source image becomes unavailable. } \label{fig2}
\end{figure}

Using a generative adversarial network (GAN)-based network for image-to-image translation has been considered an effective approach to bridge the domain gap between the source domain and target domain in some other medical imaging studies~\cite{xing2019adversarial,kamnitsas2017unsupervised}. To the best of our knowledge, there is no previous work that leverages regular fundus images to help train UWF diagnosis models. GAN-based network such as CycleGAN~\cite{zhu2017unpaired} does not require the images from source and target domain to be paired. However, considering some complex multiclassification tasks, it is assumed that the diagnosis task needs to grade the disease severity. If all the severities of the samples are mixed together for transfer training, the generated image will have difficulty defining its "Ground Truth". \cite{xing2019adversarial} used pseudo-labeling~\cite{lee2013pseudo} to annotate generated target samples, but it did not consider the wrong predictions (noise) in the pseudo-labels. The shortage of samples makes it impossible to train a GAN for each severity separately. Furthermore, in practical scenarios, we sometimes collect data from the open class, and utilizing these data is also challenging for GANs. We provide a detailed illustration and explanations of this challenge in Fig.~\ref{fig2} and Sec.~\ref{sec3.31}.

\color{black}{Therefore, we aim to develop a domain adaptation framework that can leverage a large number of regular fundus images and transfer their rich feature information to the UWF fundus images with only a small amount of data to help in UWF fundus diagnosis model training. This framework does not require the two-domain images to be paired and can be generalized to multiple tasks for the computer-aided diagnosis with UWF fundus images. In this paper, our contribution can be summarized as follows:}

\begin{enumerate}
\item For the first time, we study how to leverage regular fundus images to assist a model with learning to diagnose fundus diseases from only limited UWF imaging annotations. We transfer the regular fundus images to UWF fundus images by using a GAN-based model to introduce additional data for existing limited UWF fundus imaging samples.
\item  We propose to use the preprocessing for artifact removal in UWF fundus images from the perspective of interpretability. We maximize the use of these low-quality samples under the condition of limited training samples.
\item  We have fully studied the limitations of the GAN-based network for unpaired image transference and design a classification-based consistency regularization term for the GAN to regulate the generated image whose corresponding source image comes from a different class or even the open class distribution that does not exist in the training set.

\item Given that the original labels for generated samples are unavailable, we apply the pseudo-labeling technique and design a new loss function to more effectively extract key information from the unlabeled data. We have studied the effect of incorrect predictions in pseudo-labels on the performance of the model. Our method is proven to be significantly robust to the incorrect predictions of the pseudo-labels.
\item We evaluated our methods on the automatic diagnostic tasks of common retinal diseases, including DR grading, lesion detection and tessellated fundus segmentation. Our experimental results demonstrate that our methods can be well generalized to multiple fundus screening tasks.
\end{enumerate}

\section{Related Work}
	\subsection{Domain Adaptation}
	
	In computer vision, due to the different image characteristics of two domains (such as day and night), the performance of the classification model trained for one specific domain cannot always be guaranteed while directly being applied to other domains (domain gap~\cite{wei2018person}. To address this challenge, one approach is trying to find a common-feature space, which promises the similar features of two domains to be shared and represented, to enable co-training for cross-domain recognition problem~\cite{chen2011co,long2013transfer,peng2018cross}. The generative adversarial network (GAN)~\cite{goodfellow2014generative} is considered as another efficient approach to bridge this gap. Recent works achieve image-to-image translation on natural images~\cite{isola2017image,liu2017unsupervised,zhu2017unpaired,kim2017learning,hong2018conditional}. \cite{shaham2019singan} introduced an unconditional generative
	model that can be learned from a single image and complete multiple image processing tasks. The translated images can further be used for target domain training. 
	
	In the field of medical image analysis, domain adaptation is also an active topic for training with cross-domain datasets acquired from different types of imaging equipment. \cite{ghafoorian2017transfer} investigated the fine-tuning technique on the brain lesion segmentation application. \cite{huang2017simultaneous} studied the probability of generating high-resolution and multimodal images from low-resolution single-modality imagery using sparse coding method. With the huge domain gap still existing in some complex tasks, GANs have attracted wide interests and obvious improvements have been achieved. Mainstream works focus on the translation between CT and MRI images which are fundamentally differentt in modality and imaging principle~\cite{kamnitsas2017unsupervised,nie2017medical,dou2018unsupervised,dou2020unpaired}. \cite{kumar2019co} presented an approach for integrating PET and CT information. In microscopy image quantification, the cycle-GAN-based domain adaptation is also proved to be effective in cross-domain detection tasks~\cite{xing2019adversarial}. In the field of fundus image, \cite{costa2017towards} proposed to generate synthetic retinal images from vessels annotation. Our previous work~\cite{ju2020bridge} studied applying CycleGAN~\cite{zhu2017unpaired} with consistency regularization~\cite{zhang2019consistency} for translating the traditional fundus image to unpaired UWF domain for the first time. The experimental results demonstrate
	that the proposed method can well adapt and transfer the knowledge from traditional fundus images to UWF fundus images and improve the performance of retinal disease recognition. Besides, there is no other work that considers to achieve the domain adaptation between regular fundus images and UWF fundus images for UWF diagnosis. We still find that there is room for further improvement in both image translating and training methods, which is expected to be generalized to more tasks of UWF fundus disease recognition. 
	
	\subsection{Semi-supervised Learning}
	Pseudo-labeling~\cite{lee2013pseudo} provides a simple and effective labeling strategy for unlabeled data and is considered to be an important technique of semi-supervised learning, but it does not consider how to deal with the noise from the wrong predictions. Inspired by unsupervised learning and representation learning, \cite{rasmus2015semi} proposed a ladder network-based semi-supervised learning framework to achieve information fusion from the supervised part and unsupervised part. \cite{laine2016temporal} simplified the ladder network and kept the main idea of applying consistency regularization for extracting information from unlabelled data, which is regarded as a golden standard in the later semi-supervised learning works. \cite{miyato2018virtual} proved that the consistency regularization and entropy minimization regularization has a significant effect on mining information from unlabeled data to improve the generalization ability of the model. \cite{berthelot2019mixmatch} integrates the consistency regularization, pseudo-labeling technique etc., and achieves significantly better results than the previous semi-supervised learning technology. And in the case of a small amount of labelled data, it can also be effective. \cite{xie2019unsupervised} proposed UDA framework with some state-of-the-art techniques such as RandAugment~\cite{cubuk2019randaugment}, and uses Training Signal Annealing (TSA) to balance the supervised signal and unsupervised signal. UDA achieves excellent performance both on text and image classification tasks. Besides, SSL is also considered to be effective on more tasks, such as objects detection and semantic segmentation with a combination of domain adaptation\cite{xing2019adversarial,zheng2020rectifying}.
	
\section{Datasets and Problem Overview}
\begin{figure}
	\includegraphics[width=9cm]{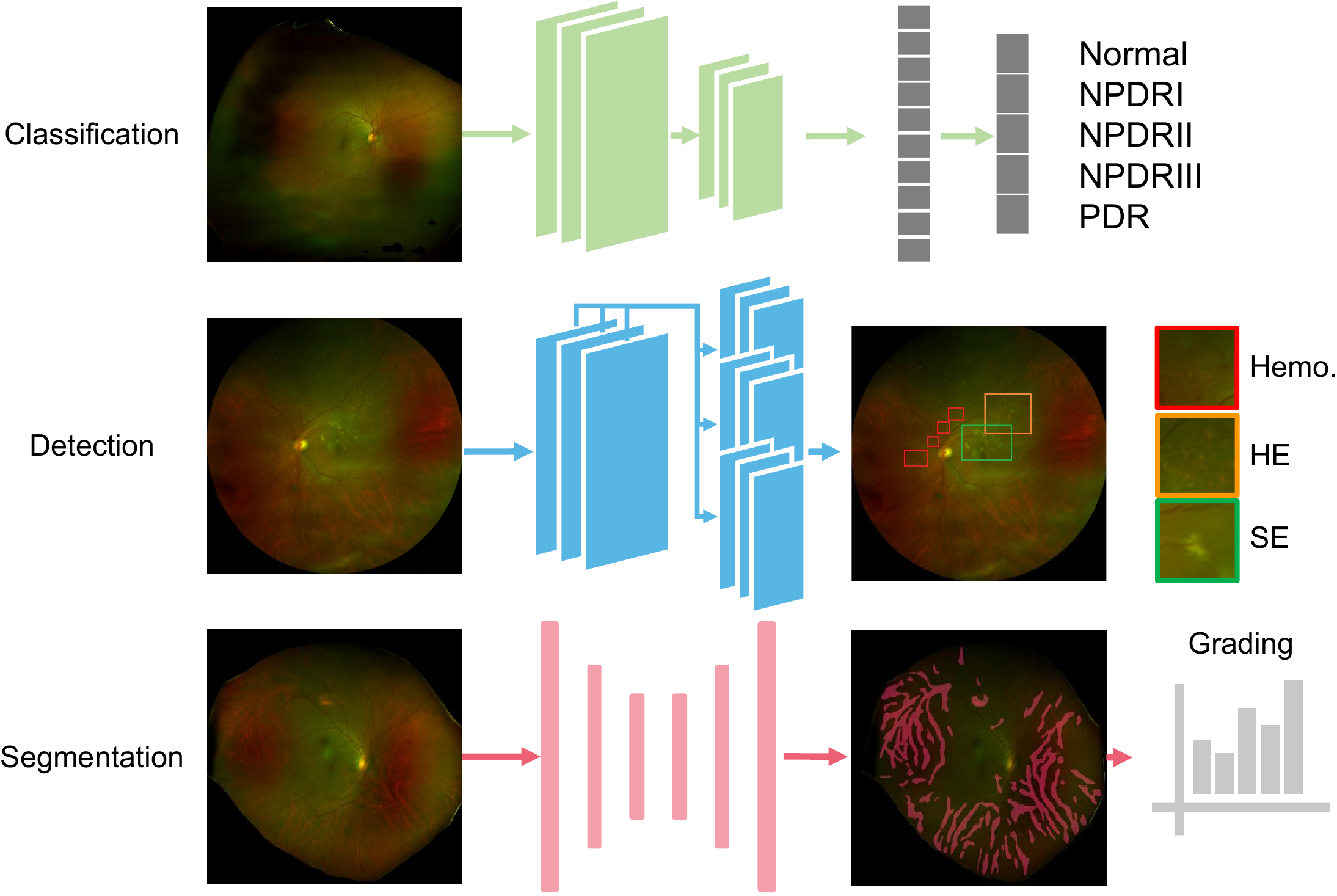}
	\centering
\caption{The overall illustration of three target tasks, multi-label classification, detection and segmentation.} \label{fig3}
\end{figure}

Our two dataset domains (the regular fundus images as the source domain and the UWF fundus images as the target domain) were acquired from private hospitals, and each image was labeled by three ophthalmologists. An image is retained only if at least two ophthalmologists are in agreement with the disease label. To further test the universality and the generalizability of our proposed methods, we compile a database that consists of different annotations for multiple tasks, i.e., classification, detection and segmentation. From the perspective of clinical application, UWF fundus shows a greater ability to cover the retina. Hence, we select several common fundus diseases diagnosis task to address the challenge as the Fig.~\ref{fig3} shows. The statistics of the datasets are shown in Table~\ref{table_1}. We divide each database into training (60\%), validation (15\%) and testing(25\%) sets with the same data distribution.

\subsection{Classification}

The grading of DR, which is the most common challenge for fundus images, is chosen to be the classification task. For this task, UWF fundus imaging with a wider field has a greater advantage because the severity level grading of DR are given according to the number of corresponding lesions in the four quadrants centered by the disc~\cite{centers2011national}. Two datasets have been built to verify our domain adaptation method, namely regular fundus dataset and UWF fundus dataset. Images from the two datasets are collected from private hospitals as well as two public datasets~\cite{KagleDiabetic,DeepDRiD} (regular and UWF fundus images, respectively) for more balanced data distribution. Then all of these fundus images are graded into five classes (0-4) by experienced ophthalmologists based on the severity. These classe labels include no DR, mild, moderate, severe nonproliferative DR (NPDR) and proliferative DR (PDR).

\renewcommand\arraystretch{1.5}

\begin{table}[]
	\centering
	
	\caption{Data statistics.  }
	\setlength{\tabcolsep}{2mm}
	\begin{tabular}{lccccc}
		\hline
		Classification   & No.                    & NPDRI          & NPDRII                                  & NPDRIII                 & PDR      \\ \hline
		Regular & 500                       & 500            & 500                                     & 500                     & 500      \\
		UWF     & 164                        & 116            & 107                                     & 103                     & 62       \\ \hline
		&                           &                &                                         &                         &          \\ \cline{1-5}
		Detection        & No.                    & Hemo.     & HE                                      & \multicolumn{1}{c}{SE} &        \\ \cline{1-5}
		Regular  &           500                &        489 (3950)        &     111 (1787)                                    & \multicolumn{1}{c}{69 (275)}   &         \\
		UWF      &           64                &        87 (712)        &      83 (793)                                   & \multicolumn{1}{c}{8 (33)}   &          \\ \cline{1-5}
		&                           &                &                                         &                         &          \\ \hline
		Segmentation   & T-0                       & T-1            & T-2                                     & T-3                     & T-4      \\ \hline
		Regular   & 200                       & 200            & 200                                     & 200                     & 200      \\
		UWF      & 37                        & 31             & 28                                      & 37                      & 42       \\ \hline

	\end{tabular}
\label{table_1}
\end{table}

\subsection{Detection}
Different from the classification task which aims to obtain the semantic label of the whole fundus image, our detection task is set to locate all lesions in a fundus image related with DR grading including hemorrhages, hard exudates and soft exudates. For the detection task, we are also concerned about the probability of domain adaptation in two-domain datasets with different semantic labels but possessing similar features. Although the DR grading dataset and this dataset have many pathology features in common, they are still defined as open sets (out of distribution) to each other. It should be emphasized that there is no overlapping between the data of different tasks; that is, each image receives only one label. Eventually, we obtained 669 and 178 images with 6012 and 1538 expert-label lesions for the source and target domains, respectively. The image-level and lesion-level statistic of the two domain datasets are shown in the second part of Table~\ref{table_1}.

\subsection{Segmentation}

The object for this task is to segment the tessellation region from the whole fundus imaging and then calculate its density which is useful to grade the severity of retinochoroidal changes~\cite{yoshihara2014objective} (shown in Fig.~\ref{fig6}). The level of disease severity corresponds to the number of specific lesions. However, tessellated fundus is distributed across the whole fundus image and difficult to label it precisely by simply counting the number of lesions. Therefore, segmentation of tessellation is required to make an accurate diagnostic decision. Besides, we also expect to grade the level of tessellated fundus for many other tasks. In this work, we divide the tessellated fundus images into 5 different grades (0-4) according to the density. For example, if the tessellated density of an image exceeds that of 80\% of the samples, we judge it as the T-4 level. So we have the grade $G$ for each sample: 
\begin{equation}
    G=k \, |\,  c_{k}<\frac{AoT}{AoF}<c_{k+1},k\in [0,4]
\end{equation}

where $AoT$ and $AoF$ denote the areas of tessellated fundus and the whole fundus respectively, and $c_{k}$ denotes the density of grading $k_{th}$ we set.

\section{Methods}
\begin{figure*}
    \includegraphics[width=18.5cm]{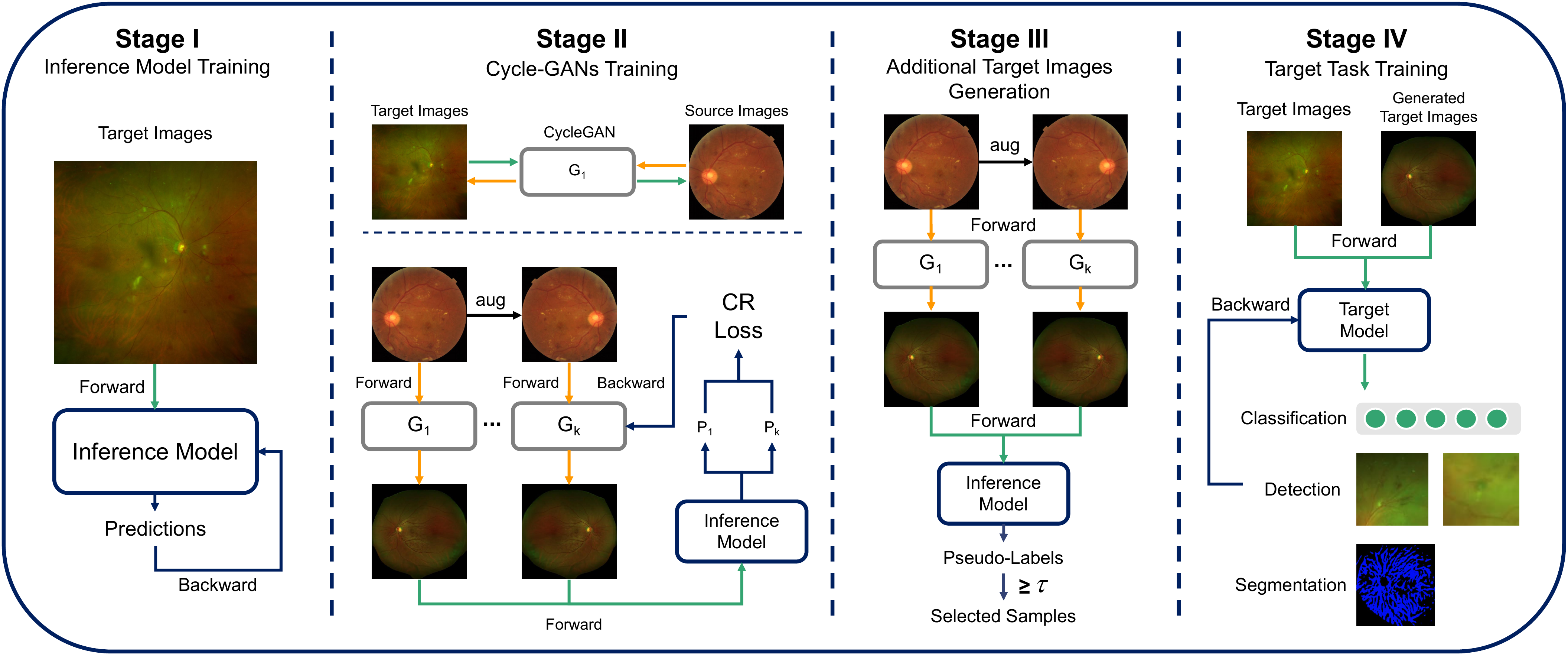}
    \centering
\caption{Overview of our proposed framework. We can divide the framework into four stages. At Stage I, we train an inference model which to be used for the later 2 stages using the existing target images. At Stage II, we train the modified CycleGAN with consistency regularization item (bottom part of the sub-figure). At Stage III, we use the Source $\rightarrow$ Target generators to pool additional generated target images and leverage the pre-trained inference model to generate the pseudo-labels. After setting a threshold for sample selection, the existing target images and additional target images with pseudo-label are trained together for three different tasks, classification, detection and segmentation respectively.} \label{fig4}
\end{figure*}

In this section, we introduce our proposed adversarial domain adaptation framework for transferring the information from regular fundus images to UWF fundus images with annotation-efficient learning. Our proposed framework can perform domain adaptation between two-domain images with different semantic labels by using a consistency regularization term for Cycle-GAN. This framework is significantly robust to the wrong predictions in pseudo-labels and can be well generalized to multiple tasks such as disease grading, lesion detection and tessellated segmentation. The overview of the proposed framework is shown in Fig.~\ref{fig4} and we divided it into four stages. At Stage I, we use the existing target images to train a target-task model, which is used to help regulate the quality of generated target samples at the Stage II and then generate pseudo-labels for generated target samples at the Stage III. At Stage IV, we use the existing samples and additional generated samples to train the target-task model together.

In Sec.~\ref{sec3.1}, we describe the necessity of preprocessing for artifact removal. In Sec.~\ref{sec3.2}, we propose using consistency regularization to regulate the quality of generated samples and achieve adversarial data augmentation. In Sec.~\ref{sec3.3}, we apply pseudo-labeling to the generated samples and design a semi-supervised loss function that learns from labeled and unlabeled data.

\subsection{Problem Definition}

Let $X^{S} = \{x_{1}^{S},x_{2}^{S},...,x_{N_{S}}^{S}\}$ and $X^{T} = \{x_{1}^{T},x_{2}^{T},...,x_{N_{T}}^{T}\}$ denote the source domain (regular fundus) and target domain (UWF fundus) images, respectively. The labels are defined as $Y^{S} = \{y_{1}^{S},y_{2}^{S},...,y_{N_{S}}^{S}\}$ and $Y^{T} = \{y_{1}^{T},y_{2}^{T},...,y_{N_{T}}^{T}\}$, respectively.
The goal is to map the regular fundus images $X^{S}$ into target domain UWF images. The generated UWF images are called pseudo-target samples $\hat{X}^{T}$, 
which are used to assist in training for UWF fundus diagnosis.

\subsection{Artifact Removal for Low-Quality Image Recycling}
\label{sec3.1}
The presence of artifacts such as eyelids and eyelashes has become a significant hurdle for making reliable retinal disease diagnoses~\cite{nagiel2016ultra}. For low-quality images, artifacts may cover more than half of the area. Even though most of the fundus  is visible and does not affect artificial screening, too much irrelevant information will result in a low-accuracy model. In the case of sparse data, discharging those poor-quality images will make the model even less accurate. Hence, we propose using segmentation approaches to achieve the necessary removal of artifacts and to recycle low-quality images. 
Since medical expertise is not heavily required for labeling artifacts, we perform some manual labeling to train a U-Net segmentation network~\cite{ronneberger2015u} to achieve pixel-level artifact removal with high precision.

\subsection{Generative Adversarial Networks for Mapping}
\label{sec3.2}
We apply CycleGAN,
whose training images do not need to be paired, as our backbone network for domain adaptation. In Sec.~\ref{sec3.31}, we introduce consistency regularization in the GANs and their loss functions. In Sec.~\ref{sec3.32}, we discuss the disadvantages of unpaired image-to-image translation GANs in our scenario and investigate how consistency regularization can help mitigate these disadvantages.
\subsubsection{Consistency Regularization for GANs}
\label{sec3.31}
Consistency regularization is gradually regarded as a gold-standard technique in semi-supervised learning (SSL)~\cite{laine2016temporal,miyato2018virtual,berthelot2019mixmatch,xie2019unsupervised}. The idea is heuristic: under the condition of not destroying semantic information, the input images for training are randomly flipped, cropped or transformed by data augmentation operations, and an additional penalty term is added to the loss function. \cite{zhang2019consistency} proposed adding consistency regularization to GANs, which aims to force the discriminator to remain unchanged by arbitrary semantic-preserving perturbations and to focus more on semantic and structural changes between real and fake data.

In this work, to further preserve pathological features and regulate the quality of the generated target images during the domain-mapping stage, we aim to train a series of consistency-regulated generators. Additional generators means that we can maximize the diversity of the generated target images, which is considered a holistic augmentation technique. Data augmentation with consistency regularization is essential in most SSL methods because it is believed that it helps extract information from unlabeled data and has been proven to have a great impact on the model performance~\cite{berthelot2019mixmatch}. We define our consistency regularization term as follows:

\begin{equation}
	\begin{split}
		r &= Randint([1, k)) \, \,  k \in [2, K], \\
		L_{cr} &=  \|( h_{T}(G^{r}_{S \rightarrow T}(x^{S}, \theta_{r}), \theta_{t}) \\
		&- h_{T}(G^{k}_{S \rightarrow T}(Aug(x^{S}), \theta_{k}), \theta_{t}))\|^{2}_{2},\\
		L_{total} &= L_{S \rightarrow T} + \lambda_{cr}L_{cr},
	\end{split}
\end{equation}
where $Aug(x)$ denotes a stochastic data augmentation function. $K$ is a hyperparameter of the number of generators being trained. $G^{r}_{S \rightarrow T}(x^{S})$ is the base $source-to-target$ generator randomly picked from the trained generator pool. $G^{k}_{S \rightarrow T}(x^{S})$ is the new training generator. $\lambda_{cr}$ is a hyperparameter to control the weight between the CycleGAN loss $L_{S \rightarrow T}$ and consistency regularization term $L_{cr}$. To better leverage the consistency regularization item, for each sample, we input $x^{S}$ and $\hat{x}^{S}$ generated by $G^{r}_{S \rightarrow T}(x^{S}, \theta_{r})$ and $G^{k}_{S \rightarrow T}(\hat{x}^{S}, \theta_{k})$ into the classification network $h_{T}(x, \theta_{T})$, which is trained by existing target samples $X^{T}$ from Stage I. Since the $h_{T}$ is trained from labeled existing target samples, so we can leverage the model as inference and to provide additional annotation-related knowledge for the GANs training.

For lesion detection task, we train a multi-label classification network to optimize the generators, which is the same as the network for tessellated fundus segmentation.

\subsubsection{Towards Cross-class/task Sample Synthesis}
\label{sec3.32}

In the real world, because unlabeled data have not been filtered manually, the data will inevitably be mixed with the data from outside the category or even the data from outside the domain, which is also called an open class/set. \cite{oliver2018realistic,xie2019unsupervised,berthelot2019mixmatch} discussed the effect of out-of-category unlabeled data on the model performance, but they came to different conclusions. In our scenario, CycleGAN is an unsupervised technique that does not require the input images from the source and target domains to be paired, and thus, it is very convenient to train. In~\cite{zhu2017unpaired}, images from the two domains are usually limited to one category. In this way, CycleGAN needs to pay attention to only the specific features of images from a certain category, which makes the image-to-image translation task much easier. However, when the task becomes complex or the information from the source domain has high diversity, the CycleGAN results will be uncontrollable, which is a shortcoming.

Some visual examples of uncontrollable CycleGANs are shown in Fig.~\ref{fig6}. The left half of the figure shows an example of the translation between an apple and an orange. Assuming that the apple is the source domain and orange is the target domain, the generated image already contains most of the characteristics of the orange. However, given that this image is the input of a network to classify apples vs oranges, 
categorizing this image becomes an issue. Does it belong to the category of apple or the category of orange, or should soft encoding be used, such as [0.7,0.3] to provide a fuzzy boundary between the two categories? When training with cross-task samples (as shown in the right half of the figure), which features are kept in the generated target images also remain to be discussed.

Our proposed consistency regularization term for GANs is potentially able to address this problem. Regardless of which class the source domain data come from in CycleGAN training, we always expect that the generated data can be clustered to the characteristics of the target domain, and the prediction results (pseudo-labels) of the target task are the constraints with which to induce this information.

\subsection{Learning from Generated Samples}
\label{sec3.3}
To address the problem mentioned above of a lack of labels for uncontrolled generated samples, we apply pseudo-labeling. pseudo-labeling originally retains only the $argmax$ of the model output, which means that guessing labels for unlabeled data is hard, and it is expected to encourage the model to have low-entropy (high-confidence) predictions. One assumption is that, in a traditional classification task, there is a difference in features between classes. However, in our scenario (DR grading), there is a strong relationship between classes/levels. We believe that the distribution of pseudo-labels implies more potentially useful information, so we keep the original output of the pseudo-labels, but we will still remove some samples with low confidence by setting a threshold.

For the classification task, inspired by MixMatch~\cite{berthelot2019mixmatch}, we optimize the following standard cross-entropy and L2 loss functions for labeled data and unlabeled data, respectively:

\begin{equation}
\begin{split}
L =& L_{s}+ \lambda_{u}L_{u}\\
= &\frac{1}{B}\sum_{b=1}^{B}H(p_{b},h_{T}(x_{b},\theta_{t})) \\ 
&+ \frac{\lambda_{u}}{U}\sum_{u=1}^{U}(max(q_{u})\geq \tau )\|q_{u}-h_{T}(x_{u},\theta_{t})\| ^{2}_{2},
\end{split}
\end{equation}
where $\tau$ is a scalar hyperparameter denoting the threshold above which we retain the generated samples and $\lambda_{u}$ is a fixed scalar hyperparameter denoting the relative weight of the unlabeled loss. $p_{b}$ is the GT for existing target samples. $q_{u}$ is the pseudo-labels from inference model for generated target samples and we aim to optimize the target model $h_{T}$ with parameters $\theta_{t}$.
For the segmentation task, we use the pretrained model for inference and generate the pseudo-labels and set a threshold to filtering out those pixels with low confidence\cite{zou2018unsupervised,zheng2020rectifying}. 
For the detection task, we set a threshold to filtering those bounding boxes with low confidence in the classification of category\cite{jeong2019consistency}.  Please refer to our Experiments section and Appendix for more details.

\section{Experiments}
\label{Sec.3}

We first perform a qualitative analysis on the GAN-generated images and discuss the difference between real UWF and generated UWF fundus as the training time progress and the change of the input size of the network. Then we extensively evaluate our proposed approach on three different tasks: 1) image classification (DR grading); 2) image detection (hemorrhages, hard exudate and soft exudate detection); and 3) image segmentation (tessellated fundus segmentation). \color{black}{}
We define all the datasets described in Sec.~\ref{Sec.3} as follows: (1) the DR datasets for the classification task (\textbf{DR}); (2) the 3-lesion datasets for the detection task (\textbf{3-L}); and (3) the tessellated datasets for the segmentation task (\textbf{T}). The experiment is performed using 4 NVIDIA P100 graphic cards.

\subsection{Preprocessing and Implementation}
	Since recognizing which parts are true fundus and which parts are artifacts does not require much medical background knowledge, thus, we believe that a simple deep learning-based segmentation network using only a few annotations for artifacts can be trained easily to help achieve more accurate segmentation results in the end. 
	
	In this work, all images are resized into 1024 $\times$ 1024 pixels. We use Adadelta optimizer~\cite{zeiler2012adadelta} to update the parameters for the U-net. All the hyper-parameters are set as default: lr=1.0, rho=0.95, epsilon=None and decay=0. For comparison, in addition to 200 images for training, we also labelled 50 images for validation and testing. We train 50 epochs for each experiments setting. For detected artifact pixels, we replaced it by dark pixels with value (RGB: 0,0,0).

\begin{figure}
    \includegraphics[width=8.5cm]{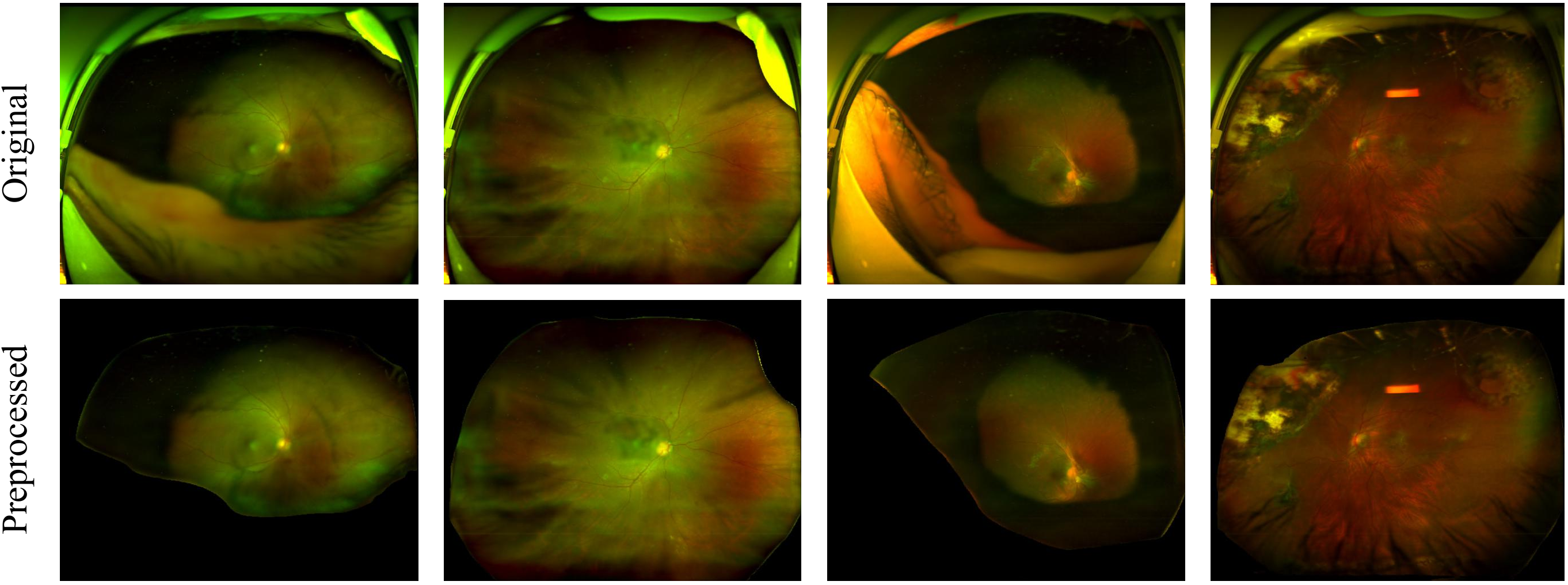}
    \centering
    \caption{Some cases of pre-processing performed by U-Net. The first row and second row denote the outcomes w/ and w/o pre-processing step.\color{black}{}} \label{fig5}
    \end{figure}

	Fig.~\ref{fig5} shows that the true fundus part can be segmented well with only a few annotations for training. See Sec.~1 of the appendix for a detailed analysis. We prove that it is a crucial and indispensable step for both generative adversarial network and UWF diagnosis model training.
	
\subsection{CycleGAN Training}
\subsubsection{Implementation Settings}
\color{black}{We resize all the images from these datasets into either 512$\times$512 or 850$\times$850 pixel images. The different sizes are chosen to test the effectiveness on the style transfer between the  two domains. For the consistency regularization term, we use $\lambda_{cr}$ = 7 and $K$ = 3 by default. We apply Adam~\cite{kingma2014adam} to optimize the generators and discriminators, where the initial learning rate is set to 0.0002 and the betas = (0.5, 0.999). We maintain a constant learning rate for the first 100 epochs and then linearly decay the learning rate to zero over the remaining epochs. We train with 150 and 300 epochs for the 512$\times$512 and 850$\times$850 pixel images, respectively.} We only apply simple flip and rotation for data augmentation while introducing consistency regularization, because we found that a too complex augmentation operation would harm to the quality of the generated target samples. There is no more improvement being observed from the performance of the diagnosis model at Stage IV either.

\subsubsection{Qualitative Generated Results Analysis}
\begin{figure}[!]
    \centering
    \includegraphics[width=9cm]{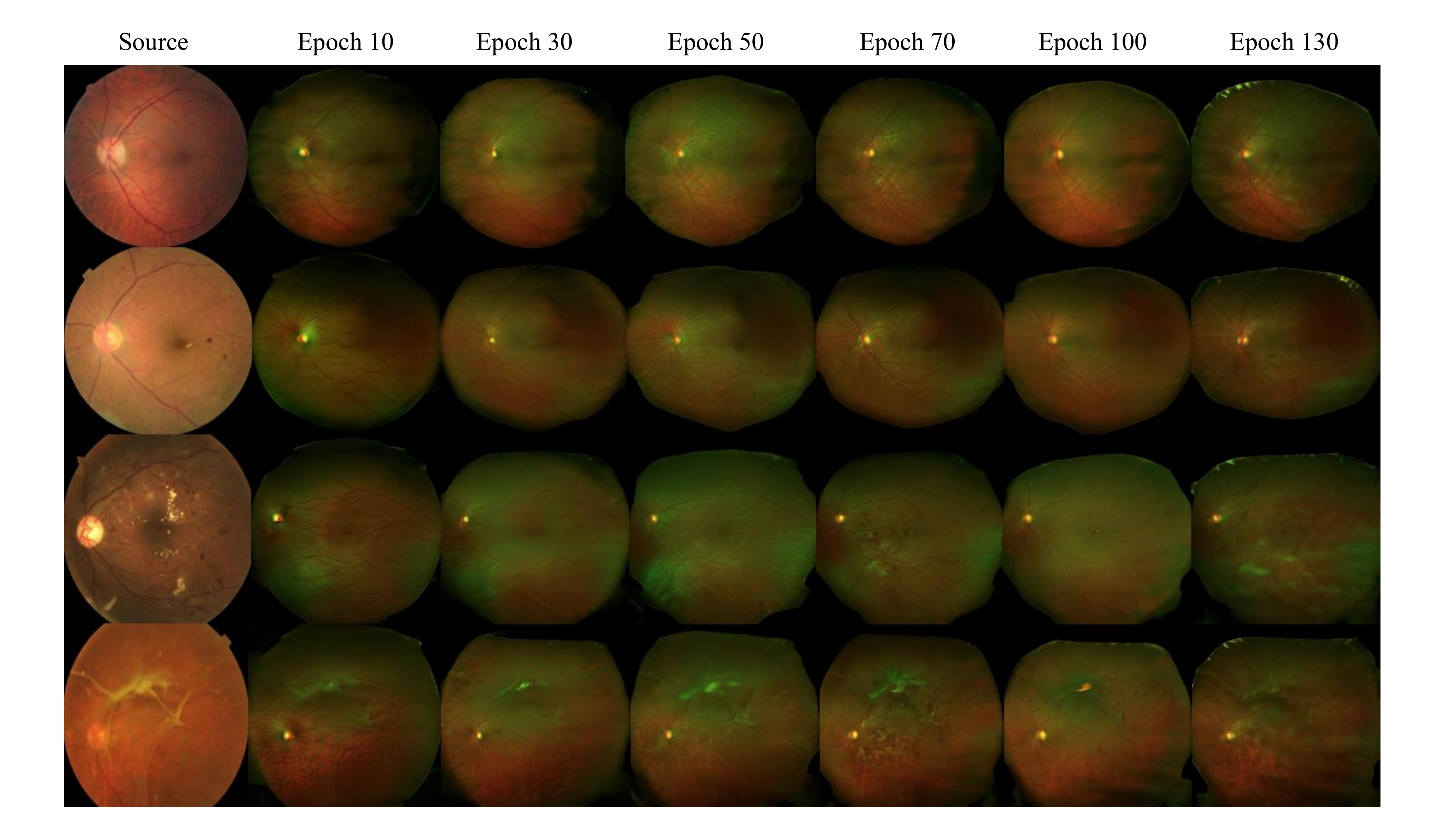}

\caption{Illustrations of the generated target images from different epochs (512 $\times$ 512). All the images are sampled from the DR task. From the top row to the bottom row, we display normal (T-4 tessellated), NPDRII (mild hemorrhages), NPDRIII (severe hemorrhages, exudates, etc.), and PDR (proliferative fibrous membrane, etc.) fundus. *Please zoom in for the best visual results.\color{black}{}}
    \label{fig6}
\end{figure}

\begin{figure}[!]
    \centering
    \includegraphics[width=8.5cm]{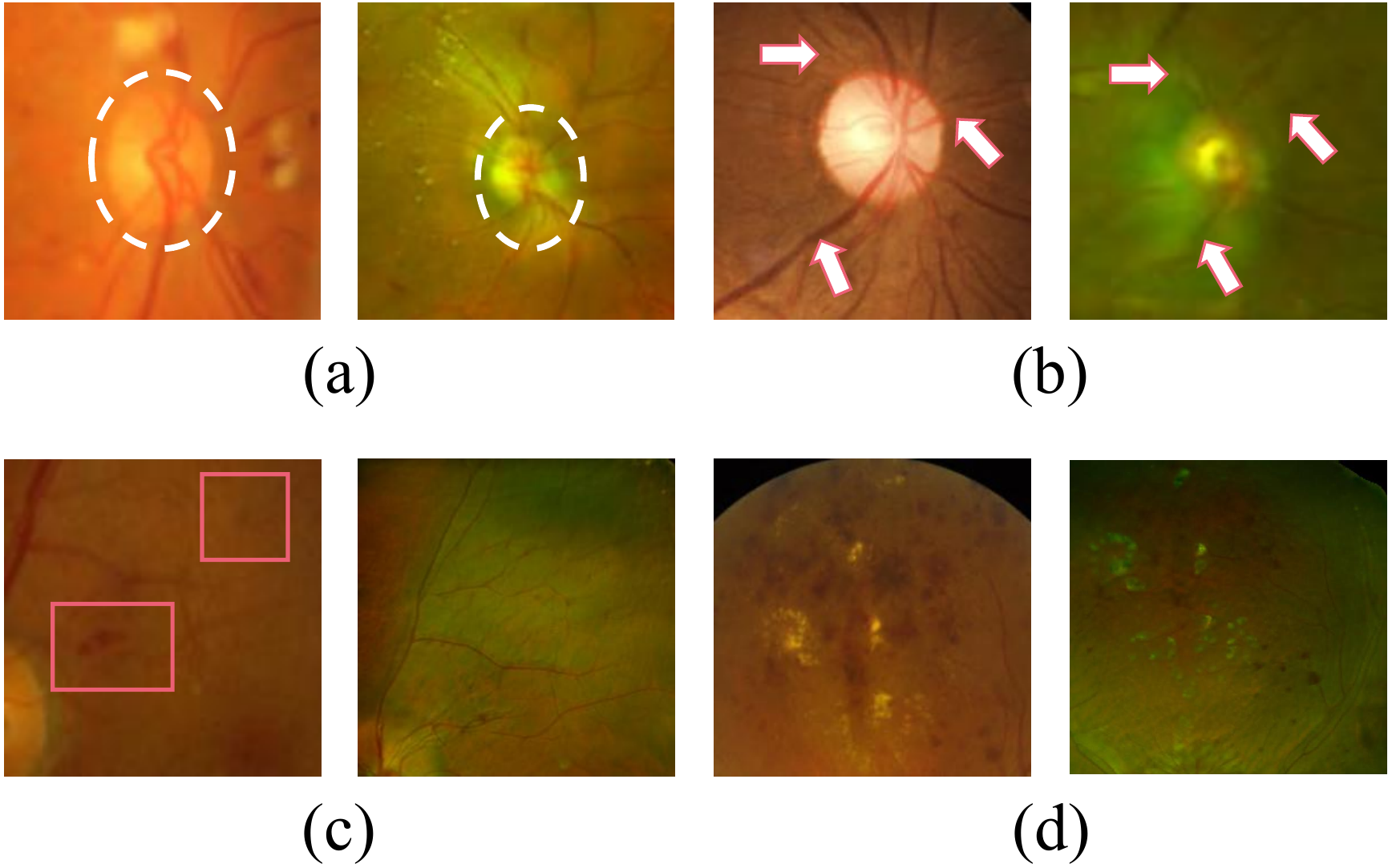}

\caption{Some examples using 850 $\times$ 850 pixel as the CycleGAN inputs size. We zoom-in to the key space for a better visualization. We can see that most lesions and pathology information are well preserved after being transferred into the target domain, however this comes with a cost of distortion of large objects such as the optic disc and blood vessels.\color{black}{}}
    \label{fig7}
\end{figure}

Our main target is to introduce additional transferred UWF fundus images to the existing UWF training images, so the most important issue is whether the lesions and their pathology features can be fully retained.
Fig.~\ref{fig6} shows the original image (with the \noindent\textbf{512} $\times$ \noindent\textbf{512} pixel input size) and its transferred UWF fundus images generated during  various training epochs. As seen from the second column of the figure, the global styling feature and components, including the relative position/ratio of the optic disc, macula, and blood vessels in the whole fundus, are fully transferable at only the 10th training epoch. Small lesions, such as hemorrhages and exudates, only start to appear at the 70th training epoch. 
However, after 100 epochs, the model tends to overfit the global features, and the fine-grained lesion features from the source domain images disappear, which means that the majority of the generated images are negative (normal) samples. Fig.~\ref{fig6} shows some examples on the translation of four different levels in DR, which also includes several kinds of lesions we mainly study in this work.  
Based on overall observations, we find that early-stopping GAN training is necessary to preserve the key lesion features from the source domain image.

Fig.~\ref{fig7} shows the generated cases when the input size is $850 \times 850$ pixels. The generated images from the $850 \times 850$ pixel input images retained more details of the key lesion features than those from the 512 $\times$ 512 pixel input images. However, we observed a distorted optic disc shape, duplicated optic discs (a) and interrupted blood vessels (b). The reason behind all these artifacts may be because the lesion features exceed the range of the perception field of the network. With higher resolution, the lesions generated (c) (d) on the pseudo-target samples are more fine-grained and close to the 
forms from the imaging of the UWF fundus camera.

To conclude, although the source domain images may do not have the ultra-wide view as the real UWF images, our model tends to map the lesions into a more global location in the generated samples, as well as the course of the vessels and tessellation, which have a simulation of both the imaging form (ratio, color etc.) and the cover of a wider range of the fundus. Small input images maintain the global features from the source images but lose fine-grained details. Large input images are able to keep key lesion features but unavoidably generate additional noise. In the following sections, we will conduct an analysis of how the quality of the generated results can affect the performance of the recognition models~\cite{gu2019progressive} in the classification, detection and segmentation tasks.
\color{black}{}

\begin{table*}[]
	\centering
	
	\caption{Classification results for diabetic retinopathy.}
	\begin{tabular}{p{1.6cm}ccclllcccccc}
		\hline
		Name/Methods     & Epochs & Source & Target & Training        & Label & Testing & Normal & NPDRI & NPDRII& NPDRIII& PDR& Avg.  \\ \hline
		Regular     & -     & -      & -      & DR-r         & GT    & DR-r & 89.12       & 88.16      & 66.56       & 79.52        & 78.56    & 80.38 \\
		UWF   & -     & -      & -      & DR-U         & GT    & DR-U & 75.00       & 51.72      & 44.44       & 67.31        & 64.52    & 58.59 \\
		Regular*   & -     & -      & -      & DR-r         & GT    & DR-U & 39.58       & 24.71      & 27.78       & 64.75        & 9.68     & 34.65 \\
		Fixed Tuning & -     & -      & -      & DR-r $\rightarrow$ DR-U  & GT    & DR-U & 96.88       & 12.07      & 29.63        & 62.14        & 16.12    & 40.09 \\
		Mix Train   & -     & -      & -      & DR-r + DR-U  & GT    & DR-U & 68.75       & 58.62      & 48.15       & 71.15        & 51.61    & 59.47 \\ \hline
		
		GAN-512     & 70    & DR-r   & DR-U   & DR-gU + DR-U & PL    & DR-U & 81.25      & 55.17      & 66.67       & 76.92        & 67.74    & \textbf{68.28} \\
		GAN-512     & 70    & DR-r   & DR-U   & DR-gU + DR-U & GT    & DR-U & 96.88      & 41.38      & 46.3       &    73.08     & 48.39   & 58.59 \\

		\hline
		GAN-850     & 300   & DR-r   & DR-U   & DR-gU + DR-U & PL    & DR-U & 
		85.00 & 71.38      & 49.63       & 82.69        & 70.32    & 70.57 \\
		GAN-850     & 300   & DR-r   & DR-U   & DR-gU + DR-U & PL*    & DR-U & 
		86.25 & 75.86   &   56.67    & 79.61        & 63.23    & \textbf{71.90} \\
		GAN-850     & 300   & DR-r   & DR-U   & DR-gU + DR-U & GT    & DR-U & 
		90.62 & 50.00   &   50.00    & 84.62        & 64.52    & 65.64 \\
		\hline
		\multicolumn{13}{l}{GAN-xxx: GAN with input size xxx; DR-r: DR regular; DR-U: DR UWF; DR-gU: DR generated UWF; GT: ground truth; PL: pseudo-labeling}  
		
	\end{tabular}
	\label{table_classi}
\end{table*}

\subsection{Classification Task}
\subsubsection{Experimental Settings}
We use a 50-layer residual neural network (ResNet-50)~\cite{he2016deep} as our classification network backbone with pretrained weights from ImageNet. We apply Adam to optimize the model. Before feeding the images to the networks, we perform extensive data augmentation. We use random flipping, random rotation. The learning rate starts at $10^{-3}$ and reduces ten-fold every 8 epochs until reaching $10^{-5}$. For the unlabeled loss, during the first 8 epochs, $\lambda_{u}$ = 1 is used for a small batch size of 8, and after that, we set $\lambda_{u}$ = 5 for a larger batch size of 16. We use the average accuracy as the evaluation metric, and all the results are reported based on 4-fold cross-validation. We keep all those settings unchanged on the inference model training at the Stage I, and target model training in the Stage IV.
\subsubsection{Overall Results}
The first section of Table~\ref{table_classi} shows the results of our baseline DR classification  methods. The first row show that, the UWF fundus classification model achieves an average accuracy of only 58.59\%.
Classifying the UWF images with the trained regular fundus image model results in an accuracy  of only 34.65\% although DR-r has more samples for training, which indicates that there is a huge domain gap between two domains. It is noted that the models perform best on the NPDRIII images, which indicates that the regular and UWF fundus images have greater similarity in terms of NPDRIII. However, the overall huge gap between two domains still exists.
We then fine-tune the pretrained regular fundus model on the UWF fundus images, but the results are still unsatisfactory, with an average accuracy of 40.09\%.
The last row demonstrates that the performance improvements of directly mixing two domains are marginal on NPDRII and NPDRIII.
The above results demonstrate that representation being learned from the regular fundus images cannot be easily generalized to the DR clarification task for UWF images even with similar pathology categories.

The second part of Table~\ref{table_classi} presents the evaluation results of the classification model after training using generated samples. When combining the transferred source domain data with pseudo-labels, the model can achieve an average accuracy of 68.28\%. However, when applying the ground-truth (GT) labels for the generated samples instead of the pseudo-labels, the average accuracy dropped dramatically from 68.28\% to 58.59\%.
It is noted that a high accuracy is achieved for the normal category due to many negative samples being generated because lesion features are lost from the source images.
Although the quality of the generated samples is still not plausible from the real UWF images, using the generated samples with pseudo-labels can improve the performance of the classification model beyond both the fine-tuning and mixed-training techniques. 

The third part of Table~\ref{table_classi} shows the CycleGAN results with the 850 $\times$ 850 pixel input images. We can observe that with a similar setting, the model trained with large images achieves an accuracy improvement from 68.28\% to 70.57\%.
In the second row (PL*), we extend our setting by using an individual threshold for each category instead of a consistent global threshold and mask out low-probability samples, which achieves an additional 
1.33 percentage point 
improvement on the classification task. This finding stems from our observation that the category with a small number of samples tends to produce smooth (low-entropy) predictive scores with low confidence scores in some categories, which we believe is not helpful for training.
The same trend from the last row can be observed when replacing the pseudo-labels with the GT labels.
\begin{figure}[!t]
	\centering
	\includegraphics[width=9cm]{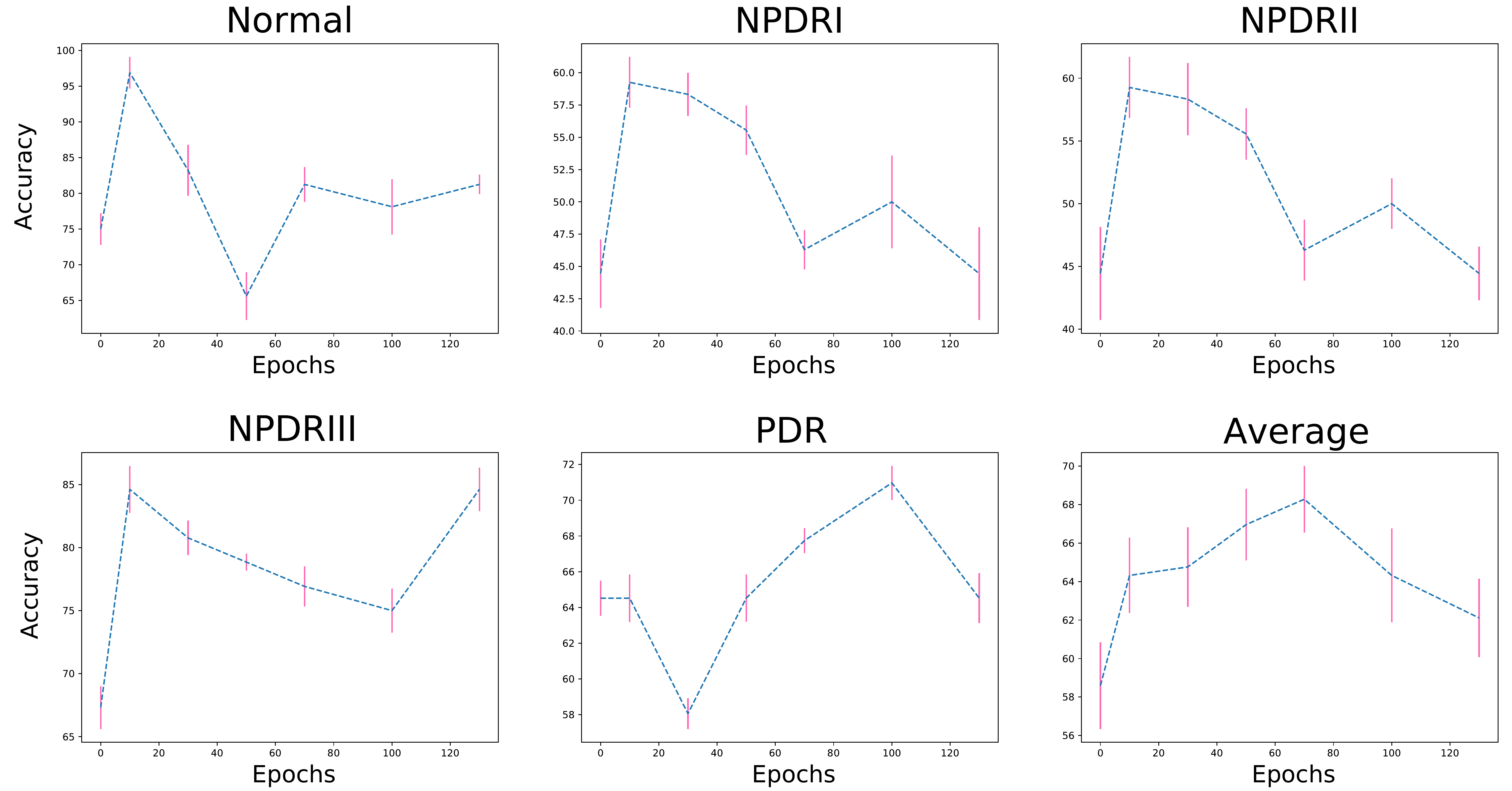}

\caption{The performance of each level of DR when using generated samples from different epochs.}
	\label{fig8}
\end{figure}

A detailed analysis of each level of DR using generated samples from different epochs is shown in Fig.~\ref{fig8}. The \noindent\textbf{Epoch 0} model is the baseline without adding any generated samples. This trend can be regarded as the distribution of the number of generated samples in different categories under different epochs. For example, in the first dozen epochs, CycleGAN only learns the general fundus style but not the lesion features. Therefore, a majority of samples are normal and NPDRI, so the classification accuracy for these two levels is high. As CycleGAN training progresses, the generated samples become diversified, and the performance for each category has been improved.

In this section, we have tried the combination of different domains and sizes of training samples, the performance is not always improved. That is to say, direct applying Cycle-GAN to generate raw pseudo-UWF fundus images (or just adding regular fundus images) are not guaranteed to improve the performance. The performance of the model is not correlated with the number of training samples but more related useful information or features. Which means more UWF/regular fundus images being collected if not engineered correctly during the model training, those extra samples without the specific knowledge constraint from target domain will not bring extra performance.

\begin{figure}[!t]
	\centering
	\includegraphics[width=9cm]{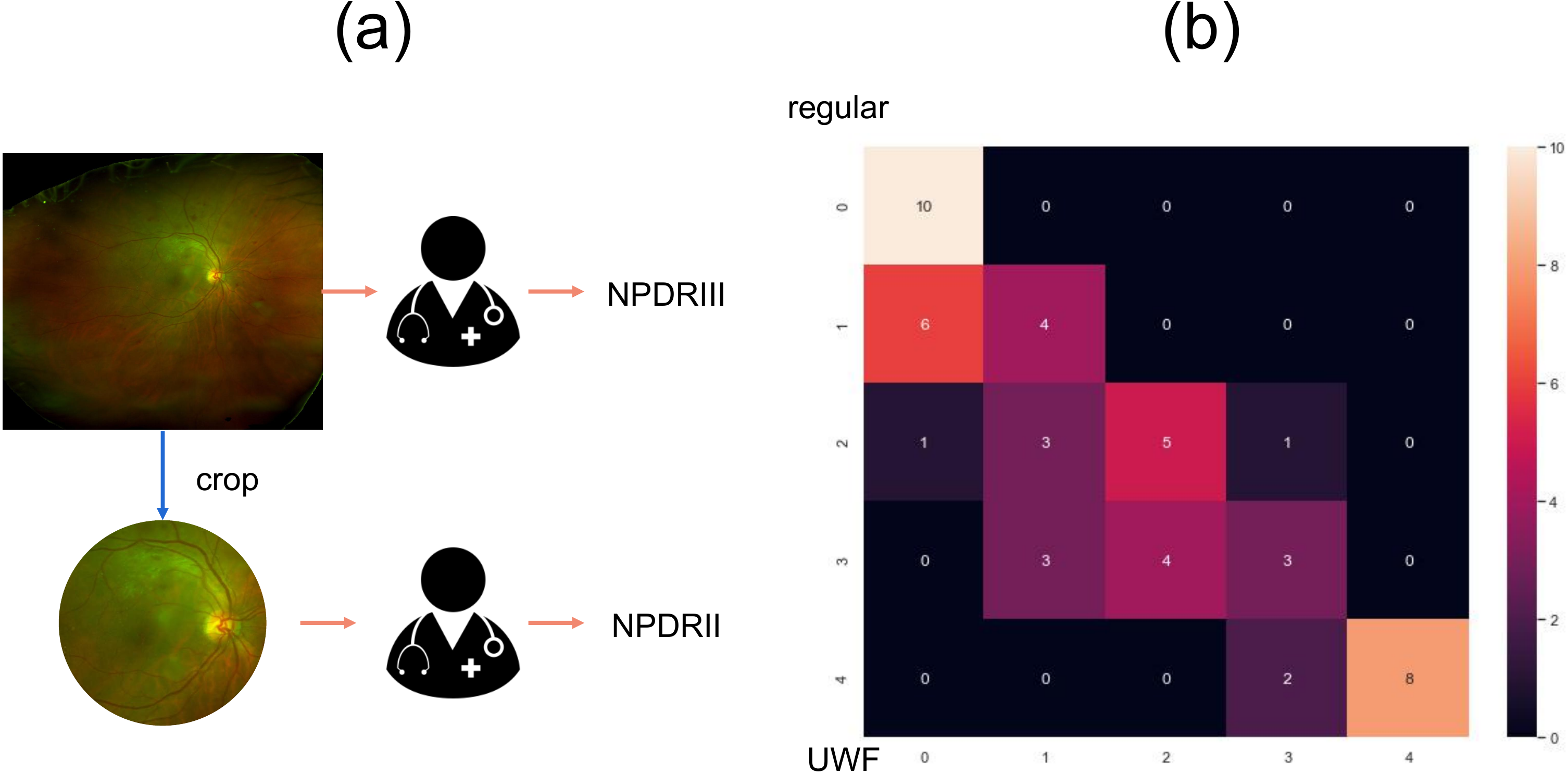}

\caption{(a) We cropped the UWF fundus images into the cover range of regular fundus images and required several ophthalmologists to relabel those cropped samples. In this case, the NPDRIII patient was wrongly diagnosed as NPDRII in the imaging scale of regular fundus imaging. (b) The confusion matrix of the real condition of patients (UWF) and diagnose results in regular fundus imaging on DR grading. Most wrong diagnoses severely affected referrals.}
	\label{fig9}
\end{figure}

\subsubsection{Discussion: Regular or UWF?} As Table~\ref{table_classi} shows, with the sufficient dataset, it seems that the performance of regular fundus imaging (80.38\%) greatly outperforms the UWF fundus imaging (baseline 58.59\% and our method 71.90\%). We have the following concerns: with computer-aided, is regular fundus more convenient or suitable for some specific retinal diseases screening such as DR since the training samples are easily to collect? To provide more evidence and direct insights for the practical value of UWF fundus imaging, we perform an extra experiment by randomly selecting 50 images from DR UWF fundus dataset and cropping them into a regular fundus range. Then, we invite three ophthalmologists to re-label those cropped fundus images, as Fig.~\ref{fig9} - (a) shows. The results are shown in Fig.~\ref{fig9} - (b). The average accuracy is only 60\%. It can be seen that there are 3 NPDRII images are wrongly diagnosed as NPDRI and 3 NPDRIII images are wrongly diagnosed as NPDRII, which seriously affect the referral. From the perspective of ERM, we treat the UWF fundus diagnosis in deep learning modelling as an upstream but also a difficult task. Due to the inconsistency annotation from diagnosis results for the same patient in two domains (such as NPDRII in regular fundus imaging but actual NPDRIII in UWF fundus imaging), so the diagnosis results can not be directly compared. Although it seems that regular fundus may have higher accuracy in DR diagnosis than UWF, it is not interpretable for the model performance. If the regular fundus imaging results are unable to accurately represent the condition of the patient, the high accuracy of the diagnosis model trained by these data may be meaningless.

\begin{minipage}[!t]{\textwidth}

\begin{minipage}[t]{0.2\textwidth}
		\centering
		\makeatletter\def\@captype{table}\makeatother\caption{Out-of-category unlabeled data.}
\footnotesize
\begin{tabular}{lll}
			\hline
\textbf{Source Data}  & \textbf{Acc.} \\ \hline
-                                       & 58.59        \\
DR                                      & 71.90        \\
3-L                               & 67.67        \\
Tess. & 62.43        \\
Random Mix                                  & 64.77        \\\hline
\end{tabular}
		\label{table_out}
\end{minipage}
\begin{minipage}[t]{0.2\textwidth}
		\centering
		\makeatletter\def\@captype{table}\makeatother\caption{Tolerance to incorrect labels.}
\footnotesize
\begin{tabular}{lll}
			\hline
\textbf{Model Name (Acc.)}  &  \textbf{Acc.} \\ \hline
-                                                                                       & 58.59                               \\
UWF (58.59\%)                                                                               & 71.90                               \\
Fixed Tuning (40.49\%)                                                                                      & 66.96                               \\
Regular* (34.65\%)                                                                                   & 64.01                               \\
Random Label                                                                                  & 65.89                               \\
			\hline
\end{tabular}
		\label{table_label}
\end{minipage}
\end{minipage}
\\
\subsubsection{Extra Categories from the Source Data}
We further extend our experiment to a real scene where the fundus images in the source domain dataset contain different diseases from the UWF fundus images in the target domain dataset.
Therefore, we set two more datasets as our source training data for domain adaptation and tests on the DR classification task: \textbf{3 Lesions} and \textbf{Tessellated}.
The results are shown in Table~\ref{table_out}\footnote{We intentionally filter out samples with low probability (threshold of 0.7) and keep high-probability samples because we think the feature distribution of those samples has a high correlation with the target tasks.}.
The first row shows the baseline. The second row illustrates the best results achieved with the DR dataset as the source domain dataset. \textbf{Random Mix} denotes that we randomly select 50\% of the images from the \textbf{3 Lesions} and \textbf{Tessellated} datasets and mix them.
The results indicate that additional categories from the source image after transferring can also improve the baseline model.
The \textbf{3 Lesions} dataset has a large overlap with the DR dataset in terms of lesions such as hemorrhages and exudates, where \textbf{Tessellated} is less relevant than the \textbf{3 Lesions} dataset.

\subsubsection{Tolerance to Incorrect Labels}
\begin{figure}[!]
	\centering
	\includegraphics[width=8cm]{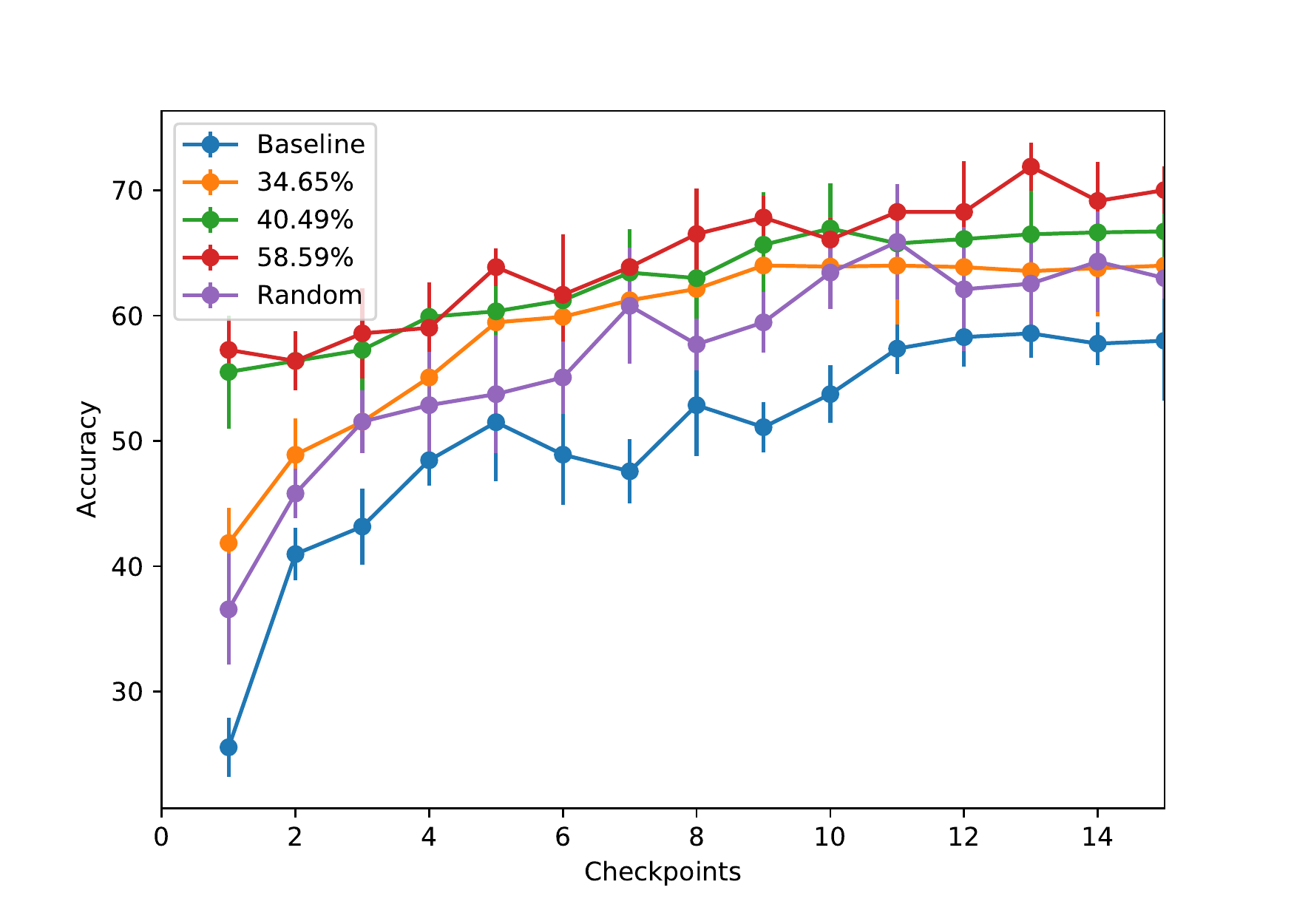}

\caption{Results of the pseudo-labels from the different inference models. The pseudo-labels generated by a model with high accuracy can help the model converge quickly, while that of a  model with low accuracy can also help improve the performance beyond the baseline.}
	\label{fig10}
\end{figure}

Furthermore, we study the ability of the proposed method to tolerate incorrect labels on the generated samples. We refer to the baseline model used to generate the pseudo-labels as the \noindent\textbf{inference model}. Because the inference model trained with limited existing UWF samples has poor performance, the majority of pseudo-label predictions are incorrect. Although we apply some common techniques used in SSL~\cite{laine2016temporal,miyato2018virtual,berthelot2019mixmatch,xie2019unsupervised} such as threshold setting, consistency regularization and unlabeled loss modules to help extract information from unlabeled data, the poor performance of the inference model in generating pseudo-labels is still a concern. To evaluate the ability our approach to tolerate incorrect labels, we use the \textbf{Regular*} and \textbf{Fixed Tuning} models from Table~\ref{table_classi} as the inference models to generate error-prone pseudo-labels for the generated target samples. The overall results are shown in Table~\ref{table_label}
and Fig.~\ref{fig10}, where the first column of the table denotes the accuracy of the inference model, which reflects the pseudo-label quality. We also apply randomly generated labels as another baseline. The results indicate that even though there are many incorrect predictions using the pseudo-label-based model, the performance can still be improved. Without any specific hyperparameter adjustment, we find that all the models outperform the baseline model. We believe that there are two reasons that can explain these results.
The first reason concerns the characteristics of the task itself. Unlike other natural image classification tasks, the disease-grading tasks show a clear correlation and progressive relationship between each category; that is, the severity is often determined by the number of lesions. Therefore, disease-grading tasks are more tolerant to incorrectly predicted labels than traditional classification tasks. Furthermore, a properly designed unsupervised regulation term can help the model slowly adapt the signal of unlabeled data rather than quickly converge to a local optimal point directed by the label. Further analysis of the consistency regularization term for its considerable resistance to incorrect labels is found in~\cite{laine2016temporal}.

\subsubsection{Comparative Study}
	\begin{table}[]
		\centering
		\caption{Comparative study results. Input size: 512 $\times$ 512.}
		\begin{tabular}{lcccc}
			\hline
			\multicolumn{4}{l}{\textbf{One-stage Adversarial Domain Adaptation}}               \\ \hline
			Methods                                                & backbone & source  & size & mAcc. \\ \hline
			Baseline                                               & ResNet-50 & - & 512  &  58.59 ($\pm$ 2.11)     \\
			IDDA \cite{zhang2019domain}         & ResNet-50 & $\checkmark$ &  512  &   50.65 ($\pm$ 6.73)    \\
			DADA \cite{dada}                     & ResNet-152 & $\checkmark$ & 512  &   46.22 ($\pm$ 7.47)    \\
			DM-ADA \cite{xu2020adversarial}      & ResNet-101 & $\checkmark$ & 512  &   38.55 ($\pm$ 5.34)      \\ \hline
			\multicolumn{4}{l}{\textbf{Two-stage Adversarial Domain Adapation}}                \\ \hline
			Methods                                                & backbone & source   & size & mAcc. \\
			Baseline                                               & ResNet-50 & -  & 512  & 58.59 ($\pm$ 2.11)      \\
			DCGAN \cite{radford2015unsupervised} & -     & $\times$     & 512  &   39.28 ($\pm$ 7.13)    \\
			SinGAN \cite{shaham2019singan}       & -     & $\times$     & 512  &   59.16 ($\pm$ 1.98)    \\
			CycleGAN \cite{zhu2017unpaired}      & -     & $\checkmark$     & 512  &   60.23 ($\pm$ 6.32)    \\
			\textbf{Ours}                                  & -   & \textbf{$\checkmark$}       & \textbf{512}  &   \textbf{68.28 ($\pm$ 4.69)}     \\ \hline
		\end{tabular}
		\label{table_com}
	\end{table}
	
	For the first time, we propose to leverage the regular fundus images for training UWF fundus diagnosis model. There is no domain adaptation study and benchmark for regular fundus and UWF fundus before. Hence, we evaluate the effectiveness of some state-of-the-art adversarial domain adaptation methods on our target task. We define two kinds of adversarial domain adaptation methods as follows:
	\begin{enumerate}
		\item \textbf{One-stage adversarial domain adaptation} is the end-to-end domain adaptation. This kind of method does not care about whether the output (mapped to the target domain) of the encoder network is close to the real target domain. That is to say, the intermediate information of adversarial domain adaptation may be difficult to understand by humans. The quality of the mapping between the two domains is completely judged by the \textbf{evaluation results on target task}. 
		
		\item \textbf{Two-stage adversarial domain adaptation} focuses on image-to-image translation between the source domain and target domain, which aims to \textbf{generate more target domain samples} to help training. The advantage of this approach is that the process of domain adaptation and synthesized results are easy to understand. Obviously, our proposed framework is a two-stage domain adaptation approach.
	\end{enumerate}
	
	We followed the official experimental settings and evaluated those methods on our datasets. All results are reported in Table~\ref{table_com}.
	
	For one-stage adversarial domain adaptation methods, we pre-train the classification model on existing labeled target samples. \textbf{Source} denotes that whether the method requires source domain images. The results of one-stage indicate that some state-of-the-art methods are difficult to be adapted to fundus images, which only obtains 50.65\% mean accuracy. This may be due to the limited fitting ability of encoder for a larger input size or more complex spatial information. In some studies, one-stage methods can be well adapted to more complex scenarios in segmentation task~\cite{pan2020unsupervised}, which is benefited from its richer information from pixel-wise annotation.
	
	For two-stage adversarial domain adaptation methods, we only change the adversarial backbone in the domain adaptation stage and keep other experimental settings unchanged. We generate the same amount of target samples for the joint training in the Stage IV. The quality of the generated target images from DCGAN \cite{radford2015unsupervised} is poor and only obtains 39.28\% mean accuracy. SinGAN~\cite{shaham2019singan} uses only one target domain image for training and the training speed is slow, and the diversity of the generated samples is also very restricted. CycleGAN \cite{zhu2017unpaired} without consistency regularization term performs better than baseline but the improvements are still limited. 
	
	In general, we further evaluate the performance of some adversarial domain adaptation approaches on DR classification problem and prove that the existing methods are difficult to be adapted to bridging the domain gap between the regular fundus and UWF fundus, and our proposed method can well address this challenge.
	
	\subsubsection{Ablation Study}
\begin{table}[]
\centering
\caption{Ablation study results. Input size: 850 $\times$ 850.}
\begin{tabular}{cccclcc}
\hline
No. & Source & CR           & $K$ & threshold  & Sep.         & Acc.  \\ \hline
1   & -      & -            & - & -          & -            & 58.59 ($\pm$ 2.11)\\
2   & DR     & $\times$     & - & -          & $\times$     & 59.47 ($\pm$ 1.01) \\
3   & DR     & $\times$     & 1 & $\times$          & $\times$     & 62.23 ($\pm$ 3.34) \\
4   & DR     & $\times$     & 3 & $\times$     & $\times$     & 60.73 ($\pm$ 4.21) \\
5   & DR     & $\checkmark$ & 3 & $\times$   & $\times$     & 64.27 ($\pm$ 5.87)\\
6   & DR     & $\checkmark$ & 3 & $\checkmark$ (uniform)     & $\times$     & 66.69 ($\pm$ 3.21) \\
7   & DR     & $\checkmark$ & 3 & $\times$   & $\checkmark$ & 67.01 ($\pm$ 5.22)\\
8   & DR     & $\checkmark$ & 3 & $\checkmark$ (uniform)     & $\checkmark$ & 70.57 ($\pm$ 1.99) \\
9   & DR     & $\checkmark$ & 3 & $\checkmark$ (individual) & $\checkmark$ & 71.90 ($\pm$ 0.90)\\ \hline
10  & 3-L    & $\times$     & 1 & $\checkmark$ (uniform)     & $\times$     & 60.02 ($\pm$ 2.06)\\
11  & 3-L    & $\checkmark$ & 3 & $\checkmark$ (individual) & $\checkmark$ & 67.67 ($\pm$ 1.99)\\ \hline
\end{tabular}
\label{table_ablation}
\end{table}
	We carry out an extensive ablation study in Table~\ref{table_ablation} to shows the effect of removing or adding components for the classification task. Specially, we measure the effect of:
	
	\begin{itemize}
	    \item Only adding the source images without domain adaptation. There is a marginal improvement from 58.59\% to 59.47\%.
	    
	    \item Leveraging non-modified CycleGAN ($K$ = 1) for domain adaptation. The performance of the model increases from 59.47\% to 62.23\%.
	    
	    \item Training several CycleGAN generators ($K$ = 3) w/o or w/ CR term for more target images generation.
	    
	    \item Using a threshold for samples selection. The $global$ denotes using an uniform threshold value and $individual$ denotes different threshold values for each sub-class.
	    
	    \item Using a newly designed loss for extracting information from labeled and unlabeled samples respectively.
	    
	    \item Using the cross-task dataset (e.g., 3-L) as source domain dataset for domain adaptation. For out-of-category source data, the model can still benefit from our proposed methods (from 60.02\% to 67.67\%).
	\end{itemize}

The ablation study results provide more and direct insights into what makes our proposed method performant.

\subsection{Detection \& Segmentation Tasks}

\begin{figure}[!]
	\centering
	\includegraphics[width=9cm]{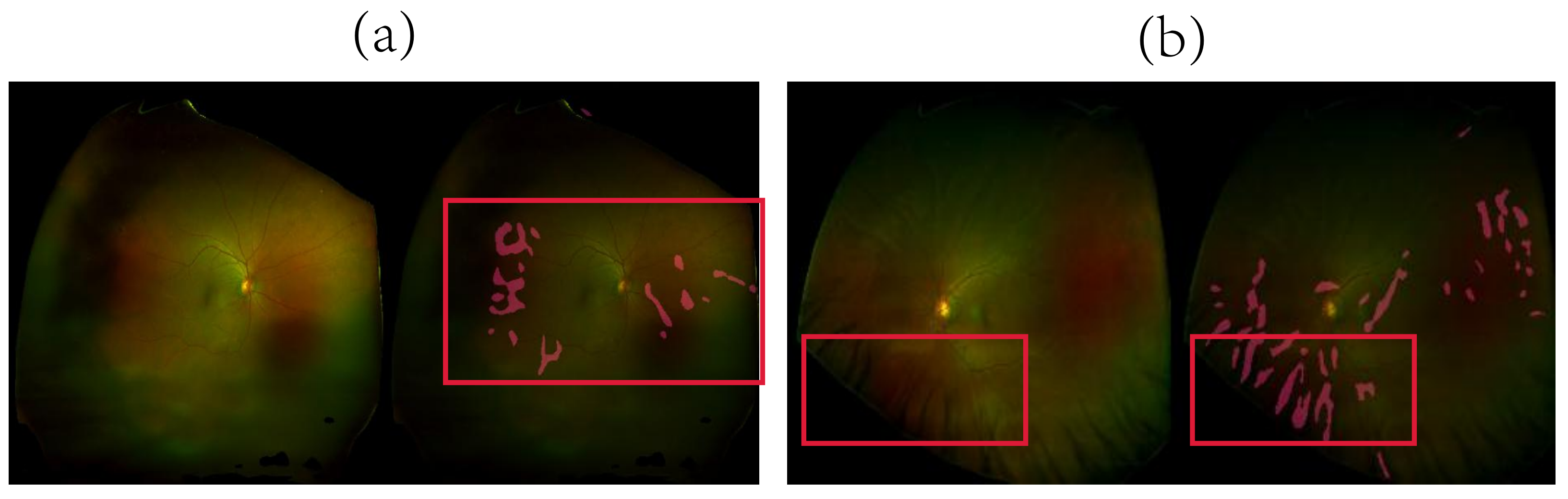}

\caption{(a) is a normal fundus, but a few tessellated areas are detected. (b) is a T-1 level tessellated fundus, but the intersection of the eyelashes and fundus is also mistakenly detected as the tessellation.}
	\label{fig11}
\end{figure}

\color{black}{Using regular fundus models to directly detect lesions and segment tessellations on UWF fundus images can achieve better results than in the classification tasks, but there will be some false positive cases. For example, in the detection task, the macula will be wrongly detected as hemorrhages; fluorescent spots near the edge of the UWF fundus can also be mistakenly detected as exudates. In the segmentation task, as Fig.~\ref{fig11} (a) shows, some normal samples will be detected as slightly tessellated fundus, where Fig.~\ref{fig11} (b) shows that the intersection of eyelashes and fundus will also be wrongly segmented as a tessellated area; however, eyelashes cannot be removed in our preprocessing work. Hence, our main goal is to reduce the number of false positives without losing the ability to detect lesions with high recall.}

\subsubsection{Detection Task}

In this work, we use YOLOv3~\cite{redmon2018yolov3} as our detection network. All the images are resized to 850 $\times$ 850 pixels. We set the nonmaximum suppression (NMS) and confidence thresholds to 0.1 and 0.3, respectively. The ignore threshold in YOLOv3 is set to 0.3. The learning rate for Adam is $10^{-5}$.
We calculate the \textbf{Precision} and \textbf{Recall} in Table~\ref{table_detect} as follow:
\begin{equation}
Precision = \frac{TP}{TP+FP}, Recall = \frac{TP}{TP+FN},
\end{equation}
and use \textbf{F1 Score} to evaluate the overall performance:
\begin{equation}
F_{1} = 2 \times \frac{ P \times R }{P+R},
\end{equation}

Our lesion detection task is very different from traditional object detection since we want to locate as many lesions as possible even if they are small in size and plain in texture. The reason is that for referral purposes, we need to achieve a high recall value while maintaining a relatively good precision.

\begin{table}[]
	\centering
	\caption{The results of the detection task.}
	\footnotesize
    \begin{tabular}{llccc}
	\hline
	\textbf{Training Data}    & \textbf{Test Data} & \textbf{a-P}              & \textbf{a-R}              & \textbf{F1}             \\ \hline
	In number of lesions    &  &              &               &             \\ \hline
	Regular                & UWF                & 5.07                      & 87.30                  & 9.58                      \\
	
	UWF (3-L) + UWF (DR) & UWF                & 5.86                      & \textbf{90.07}                    & 11.00                     \\
	gU + UWF (3-L) + UWF (DR)   & UWF                & \textbf{8.73}                      & 86.02                     & \textbf{15.85}                     \\ \hline
	In number of images    &  &              &               &             \\ \hline
	Regular                & UWF                & \multicolumn{1}{l}{30.48} & \multicolumn{1}{l}{94.88} & \multicolumn{1}{l}{46.14} \\
	
	UWF (3-L) + UWF (DR) & UWF                & \multicolumn{1}{l}{32.02} & \multicolumn{1}{l}{\textbf{95.99}} & \multicolumn{1}{l}{48.02} \\
	gU + UWF (3-L) + UWF (DR)   & UWF                & \multicolumn{1}{l}{\textbf{52.11}} & \multicolumn{1}{l}{91.01} & \multicolumn{1}{l}{\textbf{66.27}} \\ \hline
	\multicolumn{5}{l}{*a-P: average precision; a-R: average recall; F1: F1 score}     \\
	  
	\end{tabular}
	\label{table_detect}
\end{table}

We show the performance results of the detection task in Table~\ref{table_detect} with two evaluation metrics, which are lesion- and image-level-based metrics.
When the lesion-level-based metric is used as the detection statistic, the overall precision is lower than with the image-level-based metric due to the detection of too many artifacts, unlabeled and small, unknown lesions.
Considering that the DR and 3-L datasets have similar feature spaces, we add these samples in the training step for data augmentation purposes. For the pseudo-labels of the detection box on those unlabeled samples, we use 0.45 as the confidence threshold for the sample filter. For the normal images, we do not apply pseudo-labeling. We fine-tune the regular fundus model on the newly added samples. \noindent\textbf{Training Data} denotes which newly added samples we use.

The first row of Table~\ref{table_detect} is our baseline. The second row shows that there are overall improvements in the precision, recall and F1 score when extra transferred images are introduced. An extra number of positive training samples can significantly improve the performance of the model, especially when we set a relatively high threshold for the pseudo-labels. However, the results show that the problem of a large number of false positives is not solved since the precision is still low. When applying the same method when adding the generated image (confidence threshold of 0.32), the precision was significantly improved 
from 32.02\% to 52.11\% in terms of the number of images. We find that as a nongeneral feature, artifacts hardly exist in the first 90\% of the epochs. In addition, the majority of generated images are used as normal samples after the threshold is set, which will reduce the recall. Although in the detection of the lesions, we usually expect the recall to be as high as possible since we do not want to misreport any potential risk to patients, we also do not want the model to always incorrectly assume a normal sample as a disease or even a serious disease. Our methods achieved 
21.63 and 20.13 percentage point improvements on the average precision and F1 score, 
respectively, and the high generalization ability of our method in the detection task is proven.

\begin{figure*}[!h]
		\centering
		\includegraphics[width=17cm]{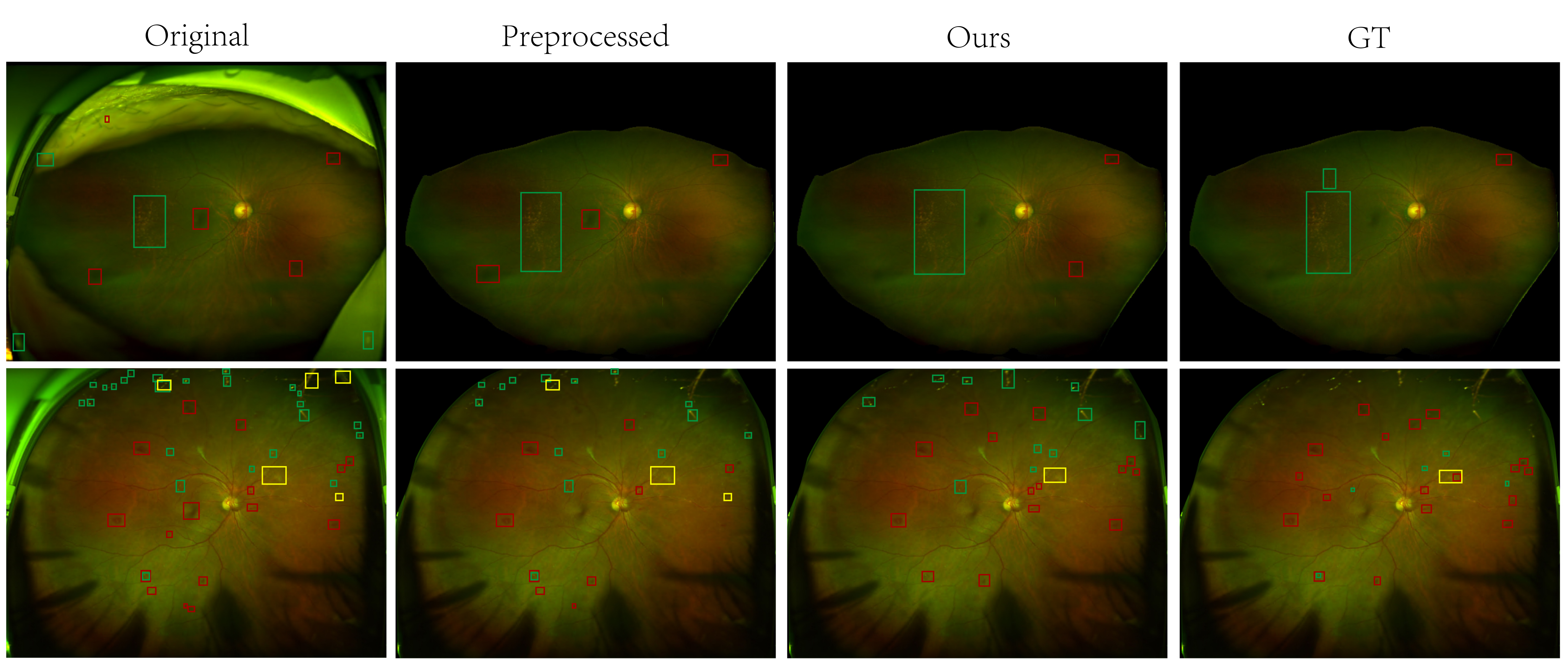}
		
		\caption{The results of the detection task. (red box: haemorrhage; green box: hard exudates; yellow box: soft exudates.) The first column - Original denotes that some artifacts can be easily detected as lesions without preprocessing. After preprocessed, there are still some false-positive cases. The third column shows that our proposed method can well reduce the false-positive cases without losing too much precision.}
		\label{fig12}
	\end{figure*}
Fig.~\ref{fig12} shows the visualization of detection task results. The original column indicates that a lot of false-positive objects are detected on the artifacts and macular is detected as haemorrhage. In some easy cases (the first row), the false detection on these non-fundus parts can be removed after preprocessing. However, there are still some hard cases of artifacts that cannot be removed by preprocessing (the second row). Our method reduces most false-positive results by adding generated samples and improves the detection performance of the model. Although some lesions are distributed around the edge of the fundus, which is out of the imaging range by regular fundus imaging, our proposed method can still detect these lesions with generated target images.

\subsubsection{Segmentation Task}
For the segmentation network architectures, we use U-Net~\cite{ronneberger2015u} with a ResNet50~\cite{he2016deep} backbone. All  the images are resized to 512 $\times$ 512 pixels. We use the Adadelta optimizer~\cite{zeiler2012adadelta} to update the parameters of U-Net. The hyperparameters are set as follows: lr=1.0, rho=0.95, epsilon=None, and decay=0.

Different from traditional segmentation tasks, 
which aim to achieve excellent pixel-level image classification, 
our target is to correctly grade the level of tessellated fundus according to the density of the tessellated area predicted by the segmentator. Therefore, we use the tessellation-level-grading accuracy as our evaluation metric instead of some traditional segmentation metrics, such as the Dice score.

Similar to the classification and detection tasks, we apply pseudo-labeling to the generated UWF samples.
Apart from the baseline model for tessellated fundus segmentation, we train an auxiliary classification model\footnote{See our appendix for more details.} to weakly predict the level of the tessellated area to filter out false-positive samples. By increasing the proportion of normal samples (both generated and real) in the training samples and by fine-tuning the pretrained regular fundus model, we can greatly reduce the false-positive rate without sacrificing the recognition accuracy of other subcategories of the tessellated fundus, e.g., a normal fundus segmented as Level-1. For the existing real normal target samples, we build a corresponding annotation map with all “0” to help adding them to the training set. As for generated samples, when the tessellated areas are under 5\% of the whole fundus, we replaced the annotation map by all “0” to help reduce the false-positive.

\begin{figure}[!]
	\centering
	\includegraphics[width=9cm]{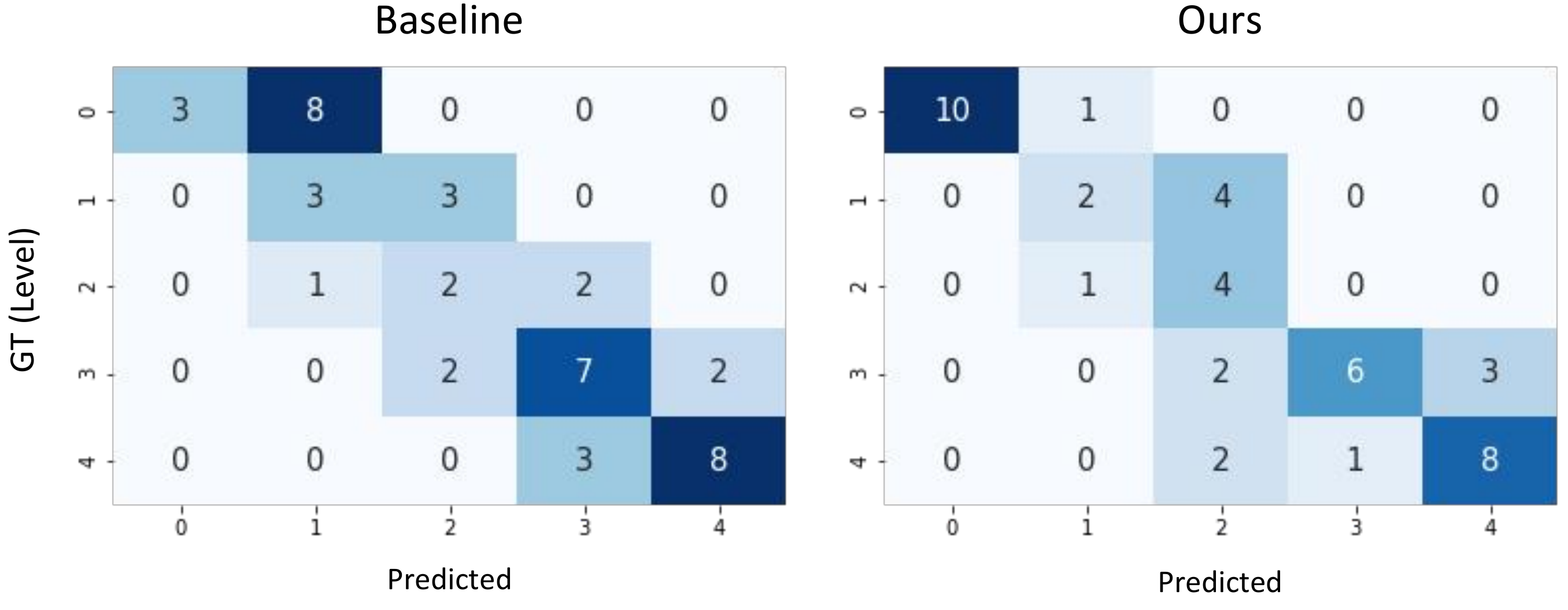}

\caption{Confusion matrix of the baseline and calibrated models for tessellated fundus grading in the segmentation task.}
	\label{fig13}
\end{figure}

We evaluate the performance of the segmentation model according to whether the predictions of the density of the tessellated areas match the level. We show the confusion matrix results in Fig.~\ref{fig13}. The left subfigure shows the segmented results using the regular fundus model, and the right part shows the results calibrated by our proposed method. We can observe that our proposed method has increased the average accuracy from 52.27\% to 68.18\%, and fewer normal samples are mistakenly segmented as Level-1.

In general, we find that for both the detection and segmentation tasks, the performance gap between the source and target domains is not as large as in the classification task. This finding may be due to the stronger and richer semantic information that can be learned by detailed annotations from the detection and segmentation tasks. Therefore, the gap between the two domains can be minimized by adding a small number of generated images to calibrate the model and adapt it to the target domain.

\subsection{Discussion on Consistency Regularization for GANs}
\color{black}{}
\cite{zhang2019consistency} proposed to add consistency regularization for GANs, which aims to enforce the discriminator to be unchanged by arbitrary semantic-preserving perturbations and to focus more on semantic and structural changes between real and fake data, and the discriminator is optimized in this work. In our proposed methods, we first use the classification model as the discriminator to build a constraint for CycleGAN training and help optimize the generator.
The benefits from consistency regularization of our training are mainly summarized as the following:
\begin{figure}[!t]
	\includegraphics[width=9cm]{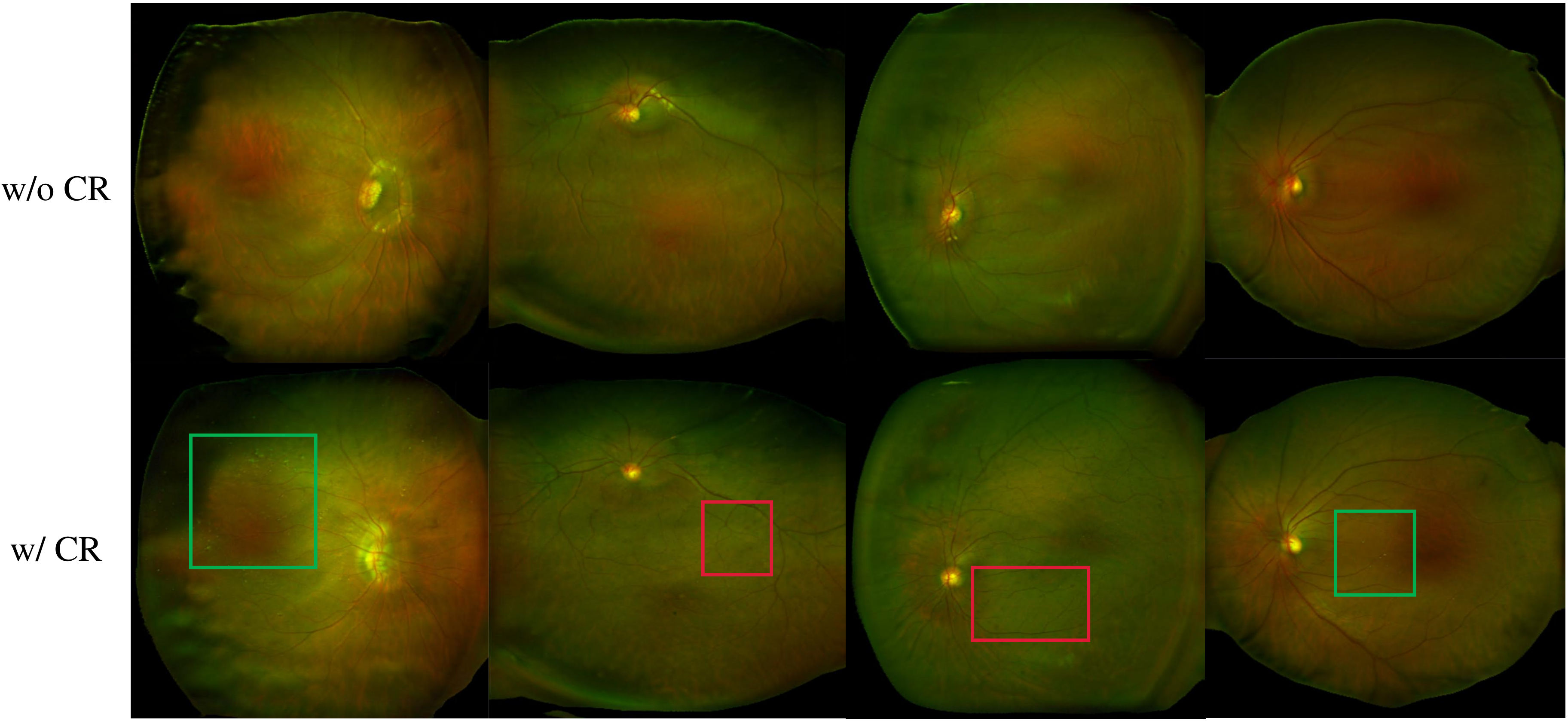}
	\centering
	\caption{The results of 300th epoch with an input size of 850 $\times$ 850. Images in the first row and second row are generated w/o and w/ consistency regularization term respectively.} \label{fig14}
\end{figure}
\begin{enumerate}
	\item MixMatch~\cite{berthelot2019mixmatch} uses consistency regularization for data augmentation on unlabelled data. This step is considered to help extract the information from unlabelled samples. In our work, the unlabelled data is generated by CycleGAN instead of real UWF fundus images. Hence, we propose to train more generators with consistency regularization term to achieve the same efficacy as these performing data augmentation.
	
	\item Rethinking towards cross-class/task samples training: Assuming that the source domain images are from out-of-distribution classes (open-class), it is unknown and uncontrollable what the generated results are. We propose to use the additional classification model with adding consistency regularization item as inference, so more additional knowledge is introduced to help train the generators in the form of knowledge distillation~\cite{hinton2015distilling}.

\end{enumerate}

In Fig.~\ref{fig14}, we give the visualization results of generated samples from the 300th epoch with cross-task datasets. The source domain is the \textbf{Dataset-Tess.} while the target domain is the \textbf{Dataset-DR}. Green box and red box denote hard exudates and hemorrhages, respectively. The first row shows results trained without consistency regularization term while the second row has it. It shows that with consistency regularization term, the generated results of the newly trained generator are closer to the target domain in the semantic feature space, although the source dataset does not contain related semantic annotation. Consistency regularization is believed to enforce the CycleGAN model to focus on important features and further regulate the quality and diversity of generated images.
\section{Conclusion}

In this paper, for the first time, we propose leveraging regular fundus images to help in training UWF fundus diagnosis models with limited UWF datasets on different tasks. We use a CycleGAN model with a consistency regularization term to transform regular fundus images to UWF fundus images. The pseudo-labeling technique is applied to introduce additional UWF images to the original training dataset. Our approach has been proven robust in multiple tasks, such as classification, detection and segmentation of UWF fundus images for diagnosis. Our work not only brings new ideas to the computer-aided diagnosis of UWF fundus but also presents an innovative methodology that enables annotation-efficient deep learning for cross-domain medical images.


%

\ifCLASSOPTIONcaptionsoff
  \newpage
\fi



%
\bibliographystyle{IEEEtran.bst}
\bibliography{tmi.bib}

\begin{thebibliography}{10}
\providecommand{\url}[1]{#1}
\csname url@samestyle\endcsname
\providecommand{\newblock}{\relax}
\providecommand{\bibinfo}[2]{#2}
\providecommand{\BIBentrySTDinterwordspacing}{\spaceskip=0pt\relax}
\providecommand{\BIBentryALTinterwordstretchfactor}{4}
\providecommand{\BIBentryALTinterwordspacing}{\spaceskip=\fontdimen2\font plus
\BIBentryALTinterwordstretchfactor\fontdimen3\font minus
  \fontdimen4\font\relax}
\providecommand{\BIBforeignlanguage}[2]{{%
\expandafter\ifx\csname l@#1\endcsname\relax
\typeout{** WARNING: IEEEtran.bst: No hyphenation pattern has been}%
\typeout{** loaded for the language `#1'. Using the pattern for}%
\typeout{** the default language instead.}%
\else
\language=\csname l@#1\endcsname
\fi
#2}}
\providecommand{\BIBdecl}{\relax}
\BIBdecl

\bibitem{fu2018joint}
H.~Fu, J.~Cheng, Y.~Xu, D.~W.~K. Wong, J.~Liu, and X.~Cao, ``Joint optic disc
  and cup segmentation based on multi-label deep network and polar
  transformation,'' \emph{IEEE transactions on medical imaging}, vol.~37,
  no.~7, pp. 1597--1605, 2018.

\bibitem{tham2014global}
Y.-C. Tham, X.~Li, T.~Y. Wong, H.~A. Quigley, T.~Aung, and C.-Y. Cheng,
  ``Global prevalence of glaucoma and projections of glaucoma burden through
  2040: a systematic review and meta-analysis,'' \emph{Ophthalmology}, vol.
  121, no.~11, pp. 2081--2090, 2014.

\bibitem{kobrin2007overview}
B.~E. Kobrin~Klein, ``Overview of epidemiologic studies of diabetic
  retinopathy,'' \emph{Ophthalmic epidemiology}, vol.~14, no.~4, pp. 179--183,
  2007.

\bibitem{kocur2002visual}
I.~Kocur and S.~Resnikoff, ``Visual impairment and blindness in europe and
  their prevention,'' \emph{British Journal of Ophthalmology}, vol.~86, no.~7,
  pp. 716--722, 2002.

\bibitem{gulshan2016development}
V.~Gulshan, L.~Peng, M.~Coram, M.~C. Stumpe, D.~Wu, A.~Narayanaswamy,
  S.~Venugopalan, K.~Widner, T.~Madams, J.~Cuadros \emph{et~al.}, ``Development
  and validation of a deep learning algorithm for detection of diabetic
  retinopathy in retinal fundus photographs,'' \emph{Jama}, vol. 316, no.~22,
  pp. 2402--2410, 2016.

\bibitem{xu2006causes}
L.~Xu, Y.~Wang, Y.~Li, Y.~Wang, T.~Cui, J.~Li, and J.~B. Jonas, ``Causes of
  blindness and visual impairment in urban and rural areas in beijing: the
  beijing eye study,'' \emph{Ophthalmology}, vol. 113, no.~7, pp. 1134--e1,
  2006.

\bibitem{nussenblatt2007age}
R.~B. Nussenblatt and F.~Ferris~III, ``Age-related macular degeneration and the
  immune response: implications for therapy,'' \emph{American journal of
  ophthalmology}, vol. 144, no.~4, pp. 618--626, 2007.

\bibitem{centers2011national}
C.~for Disease~Control, Prevention \emph{et~al.}, ``National diabetes fact
  sheet: national estimates and general information on diabetes and prediabetes
  in the united states, 2011,'' \emph{Atlanta, GA: US department of health and
  human services, centers for disease control and prevention}, vol. 201, no.~1,
  pp. 2568--2569, 2011.

\bibitem{shaw2010global}
J.~E. Shaw, R.~A. Sicree, and P.~Z. Zimmet, ``Global estimates of the
  prevalence of diabetes for 2010 and 2030,'' \emph{Diabetes research and
  clinical practice}, vol.~87, no.~1, pp. 4--14, 2010.

\bibitem{li2018efficacy}
Z.~Li, Y.~He, S.~Keel, W.~Meng, R.~T. Chang, and M.~He, ``Efficacy of a deep
  learning system for detecting glaucomatous optic neuropathy based on color
  fundus photographs,'' \emph{Ophthalmology}, vol. 125, no.~8, pp. 1199--1206,
  2018.

\bibitem{kiss2014ultra}
S.~Kiss and T.~L. Berenberg, ``Ultra widefield fundus imaging for diabetic
  retinopathy,'' \emph{Current diabetes reports}, vol.~14, no.~8, p. 514, 2014.

\bibitem{nagiel2016ultra}
A.~Nagiel, R.~A. Lalane, S.~R. Sadda, and S.~D. Schwartz, ``Ultra-widefield
  fundus imaging: a review of clinical applications and future trends,''
  \emph{Retina}, vol.~36, no.~4, pp. 660--678, 2016.

\bibitem{ding2020weakly}
L.~Ding, A.~E. Kuriyan, R.~S. Ramchandran, C.~C. Wykoff, and G.~Sharma,
  ``Weakly-supervised vessel detection in ultra-widefield fundus photography
  via iterative multi-modal registration and learning,'' \emph{IEEE
  Transactions on Medical Imaging}, 2020.

\bibitem{tan2020deep}
T.-E. Tan, D.~S.~W. Ting, T.~Y. Wong, and D.~A. Sim, ``Deep learning for
  identification of peripheral retinal degeneration using ultra-wide-field
  fundus images: is it sufficient for clinical translation?'' \emph{Annals of
  translational medicine}, vol.~8, no.~10, 2020.

\bibitem{fu2018disc}
H.~Fu, J.~Cheng, Y.~Xu, C.~Zhang, D.~W.~K. Wong, J.~Liu, and X.~Cao,
  ``Disc-aware ensemble network for glaucoma screening from fundus image,''
  \emph{IEEE transactions on medical imaging}, vol.~37, no.~11, pp. 2493--2501,
  2018.

\bibitem{jin2019dunet}
Q.~Jin, Z.~Meng, T.~D. Pham, Q.~Chen, L.~Wei, and R.~Su, ``Dunet: A deformable
  network for retinal vessel segmentation,'' \emph{Knowledge-Based Systems},
  vol. 178, pp. 149--162, 2019.

\bibitem{wang2019two}
W.~Wang, Z.~Xu, W.~Yu, J.~Zhao, J.~Yang, F.~He, Z.~Yang, D.~Chen, D.~Ding,
  Y.~Chen \emph{et~al.}, ``Two-stream cnn with loose pair training for
  multi-modal amd categorization,'' in \emph{International Conference on
  Medical Image Computing and Computer-Assisted Intervention}.\hskip 1em plus
  0.5em minus 0.4em\relax Springer, 2019, pp. 156--164.

\bibitem{wang2019retinal}
X.~Wang, L.~Ju, X.~Zhao, and Z.~Ge, ``Retinal abnormalities recognition using
  regional multitask learning,'' in \emph{International Conference on Medical
  Image Computing and Computer-Assisted Intervention}.\hskip 1em plus 0.5em
  minus 0.4em\relax Springer, 2019, pp. 30--38.

\bibitem{nagasato2018deep}
D.~Nagasato, H.~Tabuchi, H.~Ohsugi, H.~Masumoto, H.~Enno, N.~Ishitobi,
  T.~Sonobe, M.~Kameoka, M.~Niki, K.~Hayashi \emph{et~al.}, ``Deep neural
  network-based method for detecting central retinal vein occlusion using
  ultrawide-field fundus ophthalmoscopy,'' \emph{Journal of ophthalmology},
  vol. 2018, 2018.

\bibitem{ohsugi2017accuracy}
H.~Ohsugi, H.~Tabuchi, H.~Enno, and N.~Ishitobi, ``Accuracy of deep learning, a
  machine-learning technology, using ultra--wide-field fundus ophthalmoscopy
  for detecting rhegmatogenous retinal detachment,'' \emph{Scientific reports},
  vol.~7, no.~1, p. 9425, 2017.

\bibitem{levenkova2017automatic}
A.~Levenkova, A.~Sowmya, M.~Kalloniatis, A.~Ly, and A.~Ho, ``Automatic
  detection of diabetic retinopathy features in ultra-wide field retinal
  images,'' in \emph{Medical Imaging 2017: Computer-Aided Diagnosis}, vol.
  10134.\hskip 1em plus 0.5em minus 0.4em\relax International Society for
  Optics and Photonics, 2017, p. 101341M.

\bibitem{nagasawa2019accuracy}
T.~Nagasawa, H.~Tabuchi, H.~Masumoto, H.~Enno, M.~Niki, Z.~Ohara, Y.~Yoshizumi,
  H.~Ohsugi, and Y.~Mitamura, ``Accuracy of ultrawide-field fundus
  ophthalmoscopy-assisted deep learning for detecting treatment-na{\"\i}ve
  proliferative diabetic retinopathy,'' \emph{International ophthalmology},
  vol.~39, no.~10, pp. 2153--2159, 2019.

\bibitem{li2020deep}
Z.~Li, C.~Guo, D.~Nie, D.~Lin, Y.~Zhu, C.~Chen, X.~Wu, F.~Xu, C.~Jin, X.~Zhang
  \emph{et~al.}, ``Deep learning for detecting retinal detachment and
  discerning macular status using ultra-widefield fundus images,''
  \emph{Communications Biology}, vol.~3, no.~1, pp. 1--10, 2020.

\bibitem{xing2019adversarial}
F.~Xing, T.~Bennett, and D.~Ghosh, ``Adversarial domain adaptation and
  pseudo-labeling for cross-modality microscopy image quantification,'' in
  \emph{International Conference on Medical Image Computing and
  Computer-Assisted Intervention}.\hskip 1em plus 0.5em minus 0.4em\relax
  Springer, 2019, pp. 740--749.

\bibitem{kamnitsas2017unsupervised}
K.~Kamnitsas, C.~Baumgartner, C.~Ledig, V.~Newcombe, J.~Simpson, A.~Kane,
  D.~Menon, A.~Nori, A.~Criminisi, D.~Rueckert \emph{et~al.}, ``Unsupervised
  domain adaptation in brain lesion segmentation with adversarial networks,''
  in \emph{International conference on information processing in medical
  imaging}.\hskip 1em plus 0.5em minus 0.4em\relax Springer, 2017, pp.
  597--609.

\bibitem{zhu2017unpaired}
J.-Y. Zhu, T.~Park, P.~Isola, and A.~A. Efros, ``Unpaired image-to-image
  translation using cycle-consistent adversarial networks,'' in
  \emph{Proceedings of the IEEE international conference on computer vision},
  2017, pp. 2223--2232.

\bibitem{lee2013pseudo}
D.-H. Lee, ``Pseudo-label: The simple and efficient semi-supervised learning
  method for deep neural networks,'' in \emph{Workshop on Challenges in
  Representation Learning, ICML}, vol.~3, 2013, p.~2.

\bibitem{wei2018person}
L.~Wei, S.~Zhang, W.~Gao, and Q.~Tian, ``Person transfer gan to bridge domain
  gap for person re-identification,'' in \emph{Proceedings of the IEEE
  conference on computer vision and pattern recognition}, 2018, pp. 79--88.

\bibitem{chen2011co}
M.~Chen, K.~Q. Weinberger, and J.~Blitzer, ``Co-training for domain
  adaptation,'' in \emph{Advances in neural information processing systems},
  2011, pp. 2456--2464.

\bibitem{long2013transfer}
M.~Long, J.~Wang, G.~Ding, J.~Sun, and P.~S. Yu, ``Transfer feature learning
  with joint distribution adaptation,'' in \emph{Proceedings of the IEEE
  international conference on computer vision}, 2013, pp. 2200--2207.

\bibitem{peng2018cross}
M.~Peng, Q.~Zhang, Y.-g. Jiang, and X.-J. Huang, ``Cross-domain sentiment
  classification with target domain specific information,'' in
  \emph{Proceedings of the 56th Annual Meeting of the Association for
  Computational Linguistics (Volume 1: Long Papers)}, 2018, pp. 2505--2513.

\bibitem{goodfellow2014generative}
I.~Goodfellow, J.~Pouget-Abadie, M.~Mirza, B.~Xu, D.~Warde-Farley, S.~Ozair,
  A.~Courville, and Y.~Bengio, ``Generative adversarial nets,'' in
  \emph{Advances in neural information processing systems}, 2014, pp.
  2672--2680.

\bibitem{isola2017image}
P.~Isola, J.-Y. Zhu, T.~Zhou, and A.~A. Efros, ``Image-to-image translation
  with conditional adversarial networks,'' in \emph{Proceedings of the IEEE
  conference on computer vision and pattern recognition}, 2017, pp. 1125--1134.

\bibitem{liu2017unsupervised}
M.-Y. Liu, T.~Breuel, and J.~Kautz, ``Unsupervised image-to-image translation
  networks,'' in \emph{Advances in neural information processing systems},
  2017, pp. 700--708.

\bibitem{kim2017learning}
T.~Kim, M.~Cha, H.~Kim, J.~K. Lee, and J.~Kim, ``Learning to discover
  cross-domain relations with generative adversarial networks,'' in
  \emph{Proceedings of the 34th International Conference on Machine
  Learning-Volume 70}.\hskip 1em plus 0.5em minus 0.4em\relax JMLR. org, 2017,
  pp. 1857--1865.

\bibitem{hong2018conditional}
W.~Hong, Z.~Wang, M.~Yang, and J.~Yuan, ``Conditional generative adversarial
  network for structured domain adaptation,'' in \emph{Proceedings of the IEEE
  Conference on Computer Vision and Pattern Recognition}, 2018, pp. 1335--1344.

\bibitem{shaham2019singan}
T.~R. Shaham, T.~Dekel, and T.~Michaeli, ``Singan: Learning a generative model
  from a single natural image,'' in \emph{Proceedings of the IEEE International
  Conference on Computer Vision}, 2019, pp. 4570--4580.

\bibitem{ghafoorian2017transfer}
M.~Ghafoorian, A.~Mehrtash, T.~Kapur, N.~Karssemeijer, E.~Marchiori,
  M.~Pesteie, C.~R. Guttmann, F.-E. de~Leeuw, C.~M. Tempany, B.~van Ginneken
  \emph{et~al.}, ``Transfer learning for domain adaptation in mri: Application
  in brain lesion segmentation,'' in \emph{International conference on medical
  image computing and computer-assisted intervention}.\hskip 1em plus 0.5em
  minus 0.4em\relax Springer, 2017, pp. 516--524.

\bibitem{huang2017simultaneous}
Y.~Huang, L.~Shao, and A.~F. Frangi, ``Simultaneous super-resolution and
  cross-modality synthesis of 3d medical images using weakly-supervised joint
  convolutional sparse coding,'' in \emph{Proceedings of the IEEE Conference on
  Computer Vision and Pattern Recognition}, 2017, pp. 6070--6079.

\bibitem{nie2017medical}
D.~Nie, R.~Trullo, J.~Lian, C.~Petitjean, S.~Ruan, Q.~Wang, and D.~Shen,
  ``Medical image synthesis with context-aware generative adversarial
  networks,'' in \emph{International Conference on Medical Image Computing and
  Computer-Assisted Intervention}.\hskip 1em plus 0.5em minus 0.4em\relax
  Springer, 2017, pp. 417--425.

\bibitem{dou2018unsupervised}
Q.~Dou, C.~Ouyang, C.~Chen, H.~Chen, and P.-A. Heng, ``Unsupervised
  cross-modality domain adaptation of convnets for biomedical image
  segmentations with adversarial loss,'' \emph{arXiv preprint
  arXiv:1804.10916}, 2018.

\bibitem{dou2020unpaired}
Q.~Dou, Q.~Liu, P.~A. Heng, and B.~Glocker, ``Unpaired multi-modal segmentation
  via knowledge distillation,'' \emph{arXiv preprint arXiv:2001.03111}, 2020.

\bibitem{kumar2019co}
A.~Kumar, M.~Fulham, D.~Feng, and J.~Kim, ``Co-learning feature fusion maps
  from pet-ct images of lung cancer,'' \emph{IEEE Transactions on Medical
  Imaging}, vol.~39, no.~1, pp. 204--217, 2019.

\bibitem{costa2017towards}
P.~Costa, A.~Galdran, M.~I. Meyer, M.~D. Abr{\`a}moff, M.~Niemeijer, A.~M.
  Mendon{\c{c}}a, and A.~Campilho, ``Towards adversarial retinal image
  synthesis,'' \emph{arXiv preprint arXiv:1701.08974}, 2017.

\bibitem{ju2020bridge}
L.~Ju, X.~Wang, Q.~Zhou, H.~Zhu, M.~Harandi, P.~Bonnington, T.~Drummond, and
  Z.~Ge, ``Bridge the domain gap between ultra-wide-field and traditional
  fundus images via adversarial domain adaptation,'' \emph{arXiv preprint
  arXiv:2003.10042}, 2020.

\bibitem{zhang2019consistency}
H.~Zhang, Z.~Zhang, A.~Odena, and H.~Lee, ``Consistency regularization for
  generative adversarial networks,'' \emph{arXiv preprint arXiv:1910.12027},
  2019.

\bibitem{rasmus2015semi}
A.~Rasmus, M.~Berglund, M.~Honkala, H.~Valpola, and T.~Raiko, ``Semi-supervised
  learning with ladder networks,'' in \emph{Advances in neural information
  processing systems}, 2015, pp. 3546--3554.

\bibitem{laine2016temporal}
S.~Laine and T.~Aila, ``Temporal ensembling for semi-supervised learning,''
  \emph{arXiv preprint arXiv:1610.02242}, 2016.

\bibitem{miyato2018virtual}
T.~Miyato, S.-i. Maeda, M.~Koyama, and S.~Ishii, ``Virtual adversarial
  training: a regularization method for supervised and semi-supervised
  learning,'' \emph{IEEE transactions on pattern analysis and machine
  intelligence}, vol.~41, no.~8, pp. 1979--1993, 2018.

\bibitem{berthelot2019mixmatch}
D.~Berthelot, N.~Carlini, I.~Goodfellow, N.~Papernot, A.~Oliver, and C.~Raffel,
  ``Mixmatch: A holistic approach to semi-supervised learning,'' \emph{arXiv
  preprint arXiv:1905.02249}, 2019.

\bibitem{xie2019unsupervised}
Q.~Xie, Z.~Dai, E.~Hovy, M.-T. Luong, and Q.~V. Le, ``Unsupervised data
  augmentation,'' \emph{arXiv preprint arXiv:1904.12848}, 2019.

\bibitem{cubuk2019randaugment}
E.~D. Cubuk, B.~Zoph, J.~Shlens, and Q.~V. Le, ``Randaugment: Practical data
  augmentation with no separate search,'' \emph{arXiv preprint
  arXiv:1909.13719}, 2019.

\bibitem{zheng2020rectifying}
Z.~Zheng and Y.~Yang, ``Rectifying pseudo label learning via uncertainty
  estimation for domain adaptive semantic segmentation,'' \emph{arXiv preprint
  arXiv:2003.03773}, 2020.

\bibitem{KagleDiabetic}
\BIBentryALTinterwordspacing
E.~California Healthcare~Foundation. Diabetic retinopathy detection. [Online].
  Available: \url{https://www.kaggle.com/c/diabetic-retinopathy-detection}
\BIBentrySTDinterwordspacing

\bibitem{DeepDRiD}
\BIBentryALTinterwordspacing
I.~.~O. Challenge. The 2nd diabetic retinopathy – grading and image quality
  estimation challenge. [Online]. Available: \url{https://isbi.deepdr.org/}
\BIBentrySTDinterwordspacing

\bibitem{yoshihara2014objective}
N.~Yoshihara, T.~Yamashita, K.~Ohno-Matsui, and T.~Sakamoto, ``Objective
  analyses of tessellated fundi and significant correlation between degree of
  tessellation and choroidal thickness in healthy eyes,'' \emph{PloS one},
  vol.~9, no.~7, 2014.

\bibitem{ronneberger2015u}
O.~Ronneberger, P.~Fischer, and T.~Brox, ``U-net: Convolutional networks for
  biomedical image segmentation,'' in \emph{International Conference on Medical
  image computing and computer-assisted intervention}.\hskip 1em plus 0.5em
  minus 0.4em\relax Springer, 2015, pp. 234--241.

\bibitem{oliver2018realistic}
A.~Oliver, A.~Odena, C.~A. Raffel, E.~D. Cubuk, and I.~Goodfellow, ``Realistic
  evaluation of deep semi-supervised learning algorithms,'' in \emph{Advances
  in Neural Information Processing Systems}, 2018, pp. 3235--3246.

\bibitem{zou2018unsupervised}
Y.~Zou, Z.~Yu, B.~Vijaya~Kumar, and J.~Wang, ``Unsupervised domain adaptation
  for semantic segmentation via class-balanced self-training,'' in
  \emph{Proceedings of the European conference on computer vision (ECCV)},
  2018, pp. 289--305.

\bibitem{jeong2019consistency}
J.~Jeong, S.~Lee, J.~Kim, and N.~Kwak, ``Consistency-based semi-supervised
  learning for object detection,'' in \emph{Advances in neural information
  processing systems}, 2019, pp. 10\,759--10\,768.

\bibitem{zeiler2012adadelta}
M.~D. Zeiler, ``Adadelta: An adaptive learning rate method. arxiv 2012,''
  \emph{arXiv preprint arXiv:1212.5701}, vol. 1212, 2012.

\bibitem{kingma2014adam}
D.~P. Kingma and J.~Ba, ``Adam: A method for stochastic optimization,''
  \emph{arXiv preprint arXiv:1412.6980}, 2014.

\bibitem{gu2019progressive}
Y.~Gu, Z.~Ge, C.~P. Bonnington, and J.~Zhou, ``Progressive transfer learning
  and adversarial domain adaptation for cross-domain skin disease
  classification,'' \emph{IEEE Journal of Biomedical and Health Informatics},
  vol.~24, no.~5, pp. 1379--1393, 2019.

\bibitem{he2016deep}
K.~He, X.~Zhang, S.~Ren, and J.~Sun, ``Deep residual learning for image
  recognition,'' in \emph{Proceedings of the IEEE conference on computer vision
  and pattern recognition}, 2016, pp. 770--778.

\bibitem{zhang2019domain}
Y.~Zhang, H.~Tang, K.~Jia, and M.~Tan, ``Domain-symmetric networks for
  adversarial domain adaptation,'' in \emph{Proceedings of the IEEE Conference
  on Computer Vision and Pattern Recognition}, 2019, pp. 5031--5040.

\bibitem{dada}
H.~Tang and K.~Jia, ``Discriminative adversarial domain adaptation,'' in
  \emph{Association for the Advancement of Artificial Intelligence (AAAI)},
  2020.

\bibitem{xu2020adversarial}
M.~Xu, J.~Zhang, B.~Ni, T.~Li, C.~Wang, Q.~Tian, and W.~Zhang, ``Adversarial
  domain adaptation with domain mixup,'' in \emph{The Thirty-Fourth AAAI
  Conference on Artificial Intelligence}.\hskip 1em plus 0.5em minus
  0.4em\relax AAAI Press, 2020, pp. 6502--6509.

\bibitem{radford2015unsupervised}
A.~Radford, L.~Metz, and S.~Chintala, ``Unsupervised representation learning
  with deep convolutional generative adversarial networks,'' \emph{arXiv
  preprint arXiv:1511.06434}, 2015.

\bibitem{pan2020unsupervised}
F.~Pan, I.~Shin, F.~Rameau, S.~Lee, and I.~S. Kweon, ``Unsupervised
  intra-domain adaptation for semantic segmentation through self-supervision,''
  in \emph{Proceedings of the IEEE/CVF Conference on Computer Vision and
  Pattern Recognition}, 2020, pp. 3764--3773.

\bibitem{redmon2018yolov3}
J.~Redmon and A.~Farhadi, ``Yolov3: An incremental improvement,'' \emph{arXiv
  preprint arXiv:1804.02767}, 2018.

\bibitem{hinton2015distilling}
G.~Hinton, O.~Vinyals, and J.~Dean, ``Distilling the knowledge in a neural
  network,'' \emph{arXiv preprint arXiv:1503.02531}, 2015.

\end{thebibliography}

%





\section*{Supplementary material}

# Supplementary Material for "Leveraging Regular Fundus Images for Training UWF Fundus Diagnosis Models via Adversarial Learning and Pseudo-Labeling"

Lie Ju, Xin Wang, Xin Zhao, Paul Bonnington, Tom Drummond and Zongyuan Ge

*Abstract*—This is the supplementary material for the submitted manuscript "Leveraging Regular Fundus Images for Training UWF Fundus Diagnosis Models via Adversarial Learning and Pseudo-Labeling".

TABLE I
A LIST OF ANNOTATIONS.

| Notation | Explanation |
|---|---|
| $X^S$ | Regular fundus samples |
| $X^T$ | UWF fundus samples |
| $\hat{X}^T$ | Generated UWF fundus samples |
| $G_{S \to T}$ | Generators for mapping source domain to target domain |
| $K$ | The number of generators |
| $h_T$ | The inference model for generating pseudo-labels |
| $L_{cr}$ | Consistency regularization loss term for GANs |
| $\lambda_{cr}$ | Weight for $L_{cr}$ term |
| $\lambda_u$ | Weight for unlabelled data training loss function term |
| $B$ | Number of labelled data |
| $U$ | Number of unlabelled data |
| $p_b$ | Ground truth of labelled data |
| $q_u$ | Pseudo-labels of unlabelled data |
| $\tau$ | Threshold for filtering the low-confidence generated samples |

## I. PREPROCESSING

### A. Why Do We Need Preprocessing?

Different from other work ([1]), which uses preprocessing to enhance features in fundus images, we aim to remove those artifacts to enforce model to learn the semantic information we care about. artifacts are inevitable due to the UWF imaging mechanism. In addition, there is no universally accepted uniform operating standard for UWF fundus imaging yet, so in the clinical scenario, we sometimes collect some low-quality UWF fundus images. We believe that preprocessing is very important for the following considerations:

1) Previous works mostly collected their database in a controllable environment and manually abandoned those low-quality images. However, removing these data makes it worse in the condition of limited data for training.

2) The artifacts on those images may not affect the manual diagnosis of some diseases by ophthalmologists, but it has a great impact on the training of the model, especially in the case of few training data, the model will be seriously overfitting to these artifacts. We have shown a comparison of Grad-CAM visualisation results in the following experimental analysis.

For some easy cases, Otsu segmentation ([2]) is an unsupervised-based method which does not need extra annotations for artifacts to achieve quick preprocessing. However, for some hard cases, the result of Otsu segmentation can be bad, and it is also unable to quantitatively analyze the quality of segmentation results.

### B. Otsu Segmentation

artifacts like eyelids have a huge difference in colour from the fundus part, so the method of using threshold segmentation is considered an optional method. Otsu segmentation ([2]) is an unsupervised-based method which does not need extra annotations for artifacts to achieve quick preprocessing. We first locate the obstructions at the border of the image. Then we use zero value to mask out these pixels, only to keep the elliptical part from the fundus, and some segmentation results are shown in Fig. 1.

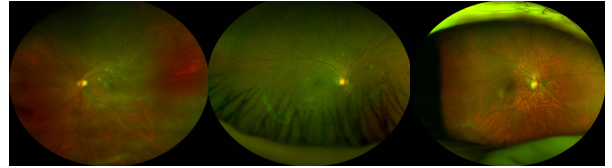

Fig. 1. Some cases of Otsu segmentation results.

We find that for most easy cases, Otsu segmentation can achieve good results and is able to alleviate the problem of overfitting to a certain extent. But there is still a certain distance from the ideal result.

### C. U-Net Segmentation

Recognizing which parts are fundus and which parts are artifacts does not require much medical background knowledge. Therefore, we believe that a simple deep learning-based segmentation network with a few annotations for artifacts can be trained to achieve more accurate segmentation results.

In this work, all images are resized into $1024 \times 1024$ pixels. We use Adadelta optimizer ([3]) to update the parameters of U-net. All the hyper-parameters are set as default: lr=1.0, rho=0.95, epsilon=None and decay=0. For comparison, in addition to the images for training, we also labelled 50 images for validation and testing. We train 50 epochs for each experiments setting.

*1) How Many Annotations Do We Need?:* Although this part of the task is relatively simple compared to disease labeling, we still hope to minimize cost. We have tested how many annotated samples we need for segmentation on artifacts. We give an illustration of the validation loss changes during the training process. As Fig. 2 shows, when the amount of training data is 100, the loss is already at a lower level, and when it increases by double, the loss does not drop more. It indicates that as a simple task, there is no need to label too much data for artifacts removal and it can attain a good segmentation performance.

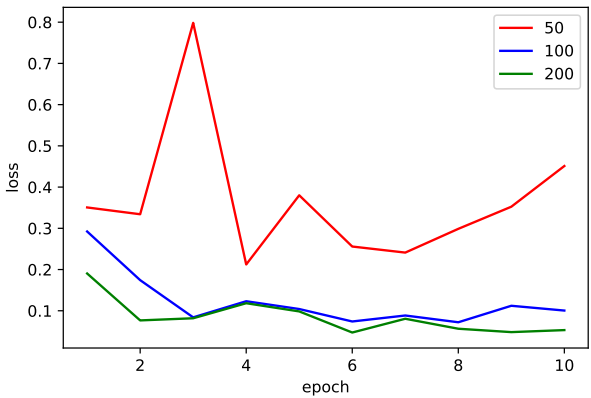

Fig. 2. Validation loss curves on the validation datasets. When the amount of data is small (50), the loss curve will fluctuate, and when the number of data increases (100 and 200), the model can converge well.

*2) U-Net Segmentation Results:* Fig. 3 shows that the fundus part can be segmented well with only a few annotations for training.

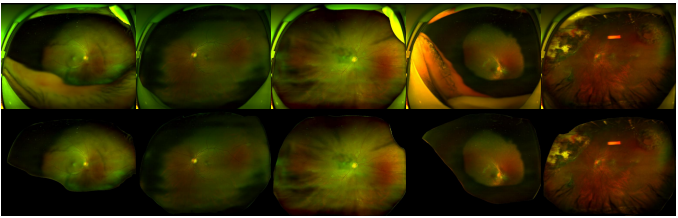

Fig. 3. Some cases of preprocessing by U-Net.

### D. Qualitative Analysis for Preprocessing

This section investigates why the classification task and GANs training both benefit a lot from the preprocessing step.

*1) Grad-CAM Visualisation:* Ophthalmologists make a diagnosis by judging whether there are related lesions in the fundus. Similar to this, the model is also classified by a similar mechanism. Gradient-weighted Class Activation Mapping (Grad-CAM) ([4]) is able to visualize the attention area of the model, which provides interpretability for the deep learning-based classification model.

We collect a tessellated fundus dataset consists of 250 normal and 250 tessellated fundus images. All image are divided into 60% for training, 20% for validation and 20% for test.

TABLE II
CLASSIFICATION RESULTS

|  | Recall | Precision | Accuracy |
| --- | --- | --- | --- |
| w/o preprocessing | 74.00 | 78.72 | 77.00 |
| w/ preprocessing | 82.00 | 85.42 | 83.00 |

Then we train two binary-classification models using images with preprocessing and without preprocessing respectively. We use InceptionV3 ([5]) as our backbone model. The input size of the model is 512 × 512. Then we select test images who have the same prediction (tessellated fundus or not) and apply the Grad-CAM technique for visualization.

We give some cases of visualization results in Fig. 4. All images before or after preprocessing are both predicted to be the same results. The first row is the original image. Obviously, although the prediction results are the same, the visualization results from images without preprocessing indicate that the model is not convincing enough. It can be seen that the attention area is mainly focused on artifacts instead of lesions. After preprocessing, the performance of the classification model has also been improved, as Table II shows.

In addition to the tessellated fundus, we also found the same issues in the evaluation of several other common retinal diseases. In other words, without preprocessing, the model can easily overfit to wrong semantic information, and it is dangerous in the clinical application.

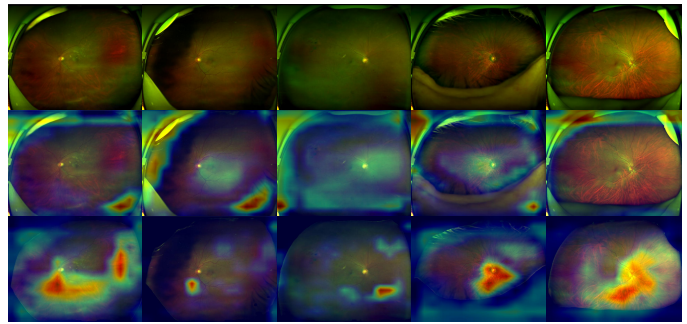

Fig. 4. Visualisation results of tessellated fundus diagnosis model. First row: original images; second row: Grad-CAM results of original images; third row: Grad-CAM results of preprocessed images.

*2) Preprocessing for CycleGAN Training:* Another effect of artifacts is CycleGAN training. As we mentioned before, the artifacts will be learned by the model as a general feature, which results in that the generated images also contain artifacts. We give some cases of CycleGAN training results in Fig. 5 with the input size of 512 × 512. Besides, the fundus part of the generated images will also be seriously affected.

## II. ADDITIONAL CLASSIFICATION TASK ANALYSIS

### A. MixUp

MixUp ([6]) is a milestone data augmentation technology to increase the training data amount. It is useful and easy to implement and widely used in recent semi-supervised learning algorithms ([7], [8]). Our previous work for 5 retinal lesions classification (haemorrhages, soft exudates, hard exudates,

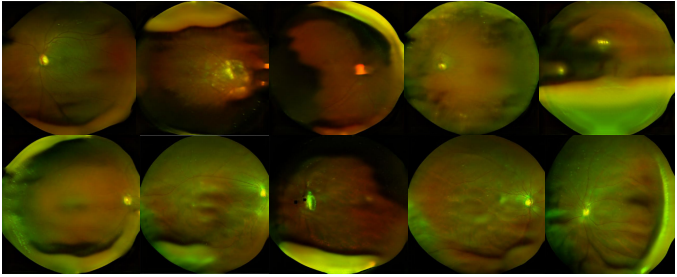

Fig. 5. Some bad cases of CycleGAN training results when using datasets without preprocessing.

drusen and retinal holes) has proved the success of MixUp on the semi-supervised learning. See [9] for more details.

However, MixUp may not work well under some specific scenarios. We give two examples in Fig. 6. The first one comes from [10]. It is a classification problem to distinguish the crack is on the right side, left side, or both. If the data in class 1 and class 2 are combined by MixUp, and the one-hot label should not be the ratio of these two but class 3. Another scenario is our DR grading task: the mix one-hot label of two NPDRII fundus images should be NPDRIII instead of NPDRII.

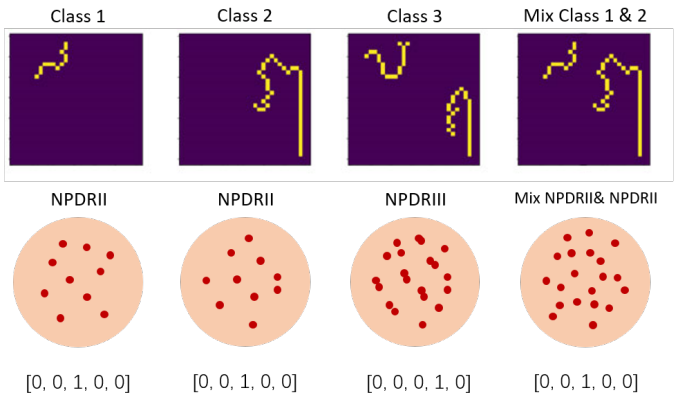

Fig. 6. Two examples show MixUp may not work well in some type of dataset.

For MixUp on detection task, please follow the implementation of GluonCV[1].

## References

[1] H. Fu, J. Cheng, Y. Xu, D. W. K. Wong, J. Liu, and X. Cao, "Joint optic disc and cup segmentation based on multi-label deep network and polar transformation," *IEEE transactions on medical imaging*, vol. 37, no. 7, pp. 1597–1605, 2018.

[2] N. Otsu, "A threshold selection method from gray-level histograms," *IEEE transactions on systems, man, and cybernetics*, vol. 9, no. 1, pp. 62–66, 1979.

[3] M. D. Zeiler, "Adadelta: An adaptive learning rate method. arxiv 2012," *arXiv preprint arXiv:1212.5701*, vol. 1212, 2012.

[4] R. R. Selvaraju, M. Cogswell, A. Das, R. Vedantam, D. Parikh, and D. Batra, "Grad-cam: Visual explanations from deep networks via gradient-based localization," in *Proceedings of the IEEE international conference on computer vision*, 2017, pp. 618–626.

[5] C. Szegedy, V. Vanhoucke, S. Ioffe, J. Shlens, and Z. Wojna, "Rethinking the inception architecture for computer vision," in *Proceedings of the IEEE conference on computer vision and pattern recognition*, 2016, pp. 2818–2826.

[6] H. Zhang, M. Cisse, Y. N. Dauphin, and D. Lopez-Paz, "mixup: Beyond empirical risk minimization," *arXiv preprint arXiv:1710.09412*, 2017.

[7] D. Berthelot, N. Carlini, E. D. Cubuk, A. Kurakin, K. Sohn, H. Zhang, and C. Raffel, "Remixmatch: Semi-supervised learning with distribution matching and augmentation anchoring," in *International Conference on Learning Representations*, 2019.

[8] D. Berthelot, N. Carlini, I. Goodfellow, N. Papernot, A. Oliver, and C. Raffel, "Mixmatch: A holistic approach to semi-supervised learning," *arXiv preprint arXiv:1905.02249*, 2019.

[9] L. Ju, X. Wang, Q. Zhou, H. Zhu, M. Harandi, P. Bonnington, T. Drummond, and Z. Ge, "Bridge the domain gap between ultra-wide-field and traditional fundus images via adversarial domain adaptation," *arXiv preprint arXiv:2003.10042*, 2020.

[10] S.-B. Yang and T.-L. Yu, "Pseudo-representation labeling semi-supervised learning," *arXiv*, pp. arXiv–2006, 2020.

 
\newpage

\end{document}